\documentclass{article}

\PassOptionsToPackage{numbers, compress}{natbib}
\usepackage[final, main]{neurips_2026}
\usepackage{booktabs}
\usepackage[table]{xcolor}
\usepackage{graphicx}
\usepackage{amsmath}
\usepackage{amssymb}
\usepackage{subfig}
\usepackage{float}
\usepackage{algorithm}
\usepackage{algpseudocode}
\algrenewcommand\algorithmicrequire{\textbf{Input:}}
\algrenewcommand\algorithmicensure{\textbf{Output:}}
\usepackage{wrapfig} 
\usepackage{subcaption}
\usepackage{tabularx}
\usepackage{array}
\usepackage[table]{xcolor}

\usepackage[utf8]{inputenc} 
\usepackage[T1]{fontenc}    
\usepackage{hyperref}       
\usepackage{cleveref} 

\usepackage{url}            
\usepackage{booktabs}       
\usepackage{amsfonts}       
\usepackage{nicefrac}       
\usepackage{microtype}      
\usepackage{xcolor}         
\usepackage{pifont}
\newcolumntype{Y}{>{\centering\arraybackslash}X}

\usepackage{soul}
\usepackage{cleveref}
\usepackage{pifont}
\usepackage{xurl}
\usepackage{xspace}
\usepackage{colortbl}
\usepackage{multirow}
\usepackage{makecell}

\usepackage{listings}

\crefname{figure}{Fig.}{Figs.}
\Crefname{figure}{Figure}{Figures}
\crefname{table}{Table}{Tables}
\Crefname{table}{Table}{Tables}

\definecolor{_green}{RGB}{0, 255, 0}
\definecolor{_red}{RGB}{237, 29, 36}
\definecolor{_blue}{RGB}{3, 113, 188}
\definecolor{_gray}{RGB}{179, 179, 179}
\definecolor{_purple}{RGB}{191, 128, 191}
\definecolor{_coral}{RGB}{255, 127, 80}
\definecolor{_orange}{RGB}{255, 165, 120}

\definecolor{_fst}{RGB}{192, 225, 201}
\definecolor{_snd}{RGB}{226, 237, 185}
\definecolor{_trd}{RGB}{255, 250, 193}

\title{iCATS: Fast Video Generation via Interaction-Aware Sparse Attention and Timestep-Adaptive Sparsity}

\author{%
  Chengfeng Han$^{1,2}$\thanks{Intern at T-Head Semiconductor Co., Ltd.} \quad
  Ang Wang$^{3}$ \quad
  Baole Ai$^{3}$ \quad
  Xianlu Bian$^{1,2}$\footnotemark[1] \quad
  Jie Yao$^{2}$ \\
  \bfseries
  Zilong Huang$^{2}$ \quad
  Dandan Ding$^{1}$\thanks{Corresponding author. \texttt{DandanDing@hznu.edu.cn}} \\[4pt]
  \mdseries\small
  $^{1}$School of Information Science and Technology, Hangzhou Normal University \\
  $^{2}$T-Head Semiconductor Co., Ltd. \quad
  $^{3}$Alibaba Group
}

\begin{document}

\maketitle

\vspace{-20pt}
\begin{figure}[H]
    \centering
    \includegraphics[width=\linewidth]{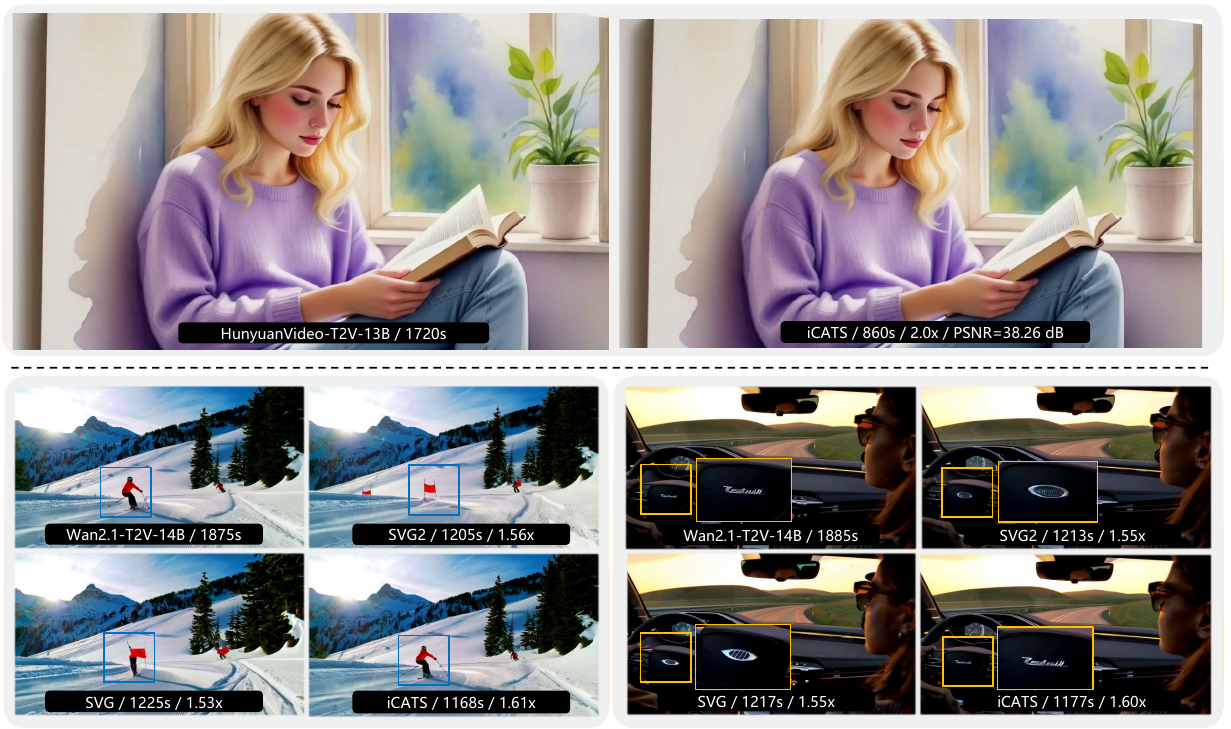}
    \caption{Visualizations on HunyuanVideo-T2V-13B and Wan2.1-T2V-14B show that iCATS better preserves visual details while achieving higher speedup than other training-free methods, producing results closer to native videos. On average, it achieves 2.03$\times$ acceleration with 31.017\,dB PSNR on HunyuanVideo-T2V-13B, and 1.55$\times$ acceleration with 29.301\,dB PSNR on Wan2.1-T2V-14B.}
    \label{fig:visual}
\end{figure}

\begin{abstract}

    Training-free sparse attention offers a practical acceleration solution to Diffusion Transformers (DiTs) via reducing computations without fine-tuning. It typically involves estimating the importance of query-key regions and deriving sparse masks to compute only the important candidates, which inevitably introduces approximation errors that may degrade generation quality.
    To better balance the efficiency-quality trade-off, we propose iCATS, integrating improved importance estimation and sparse mask construction with an efficient hardware execution strategy.
    Specifically, for importance estimation, 
    unlike previous works that perform independent clustering over query and key tokens based on feature similarity to estimate attention scores,
    iCATS demonstrates that clustering based on query-key dot-product interactions is more accurate and further reformulates this objective as a simple quadratic form for low-cost computation. For sparse mask construction, instead of using a fixed top-$p$ rule, we observe that tolerance to sparse approximation errors varies across denoising timesteps and therefore introduce an SNR-guided sparsity schedule to adjust sparsity dynamically, leading to higher accuracy. Finally, for hardware execution, we devise a tail-merging strategy to reduce padding overhead caused by irregular cluster sizes, improving GPU kernel utilization.    
    Extensive experiments show that iCATS achieves 2.03$\times$  acceleration with 31.017 dB PSNR on HunyuanVideo-T2V-13B and 1.55$\times$  acceleration with 29.301 dB PSNR on Wan2.1-T2V-14B, delivering a state-of-the-art efficiency-quality trade-off.
    \end{abstract}

\section{Introduction}
\label{sec:introduction}

Diffusion Transformers (DiTs)~\cite{Peebles2022DiT} have driven remarkable progress in video generation~\cite{wan2025, kong2024hunyuanvideo, yang2024cogvideox, gao2025seedance, seedance2026seedance20advancingvideo, klingteam2025klingomnitechnicalreport, wu2025hunyuanvideo15technicalreport, liu2024sorareviewbackgroundtechnology, feng2026quantsparse, li2026dsa, zhou2026svgearparameterfreelinearcompensation}, emerging as the dominant backbone architecture. However, as the demand for generating longer and higher-resolution videos continues to grow, the length of token sequences increases substantially, causing the quadratic computational cost of attention mechanisms to rise sharply~\cite{NIPS2017_3f5ee243}. This quadratic scaling has become a critical bottleneck during inference for DiT-based video generation models, motivating efficient attention acceleration while preserving generated video quality.

Sparse attention~\cite{zhang2025fast, chen2025rainfusion, chen2025rainfusion2, yang2025sparse, chen2025sparsevdit, li2025radial, NEURIPS2025_dfc310e8, liu2025rectified, agarwal2026monarchrtefficientattentionrealtime, luo2026trainingfreesparseattentionfast} is an effective way to accelerate DiT inference by computing only a subset of attention interactions. Existing works fall into two categories: training-based and training-free methods. Training-based methods~\cite{zhang2026sla, zhang2026sla2, chen2026sanavideo, fang2026saladachievehighsparsityattention} fine-tune models with predefined sparse patterns, achieving high sparsity (e.g., $\sim$90\%~\cite{zhang2026sla2}) but requiring retraining when sparsity changes. In contrast, training-free methods~\cite{shmilovich2025liteattentiontemporalsparseattention, NEURIPS2025_75b2a663, li2025psapyramidsparseattention, xi2025sparse, liang2026vmonarchefficientvideodiffusion, Xia_2025_ICCV} introduce sparsity without modifying model parameters. With lower sparsity (e.g., $\sim$70\%~\cite{yang2025sparse}), they provide greater flexibility and plug-and-play usability across various pretrained DiT models, which is more practical for real applications. Therefore, this work focuses on training-free sparse attention for DiT acceleration.

The core idea of training-free sparse attention is retaining the most important attention computations only during inference. This is typically achieved by first \textbf{estimating the importance} of query-key regions in the attention map and then \textbf{deriving sparse masks} through rules such as top-\(p\) selection. A common strategy is to partition query and key tokens independently and use group mean features to form a smaller query-key matrix for this importance estimation~\cite{NEURIPS2025_75b2a663, liu2025rectified, yang2025sparse}. Based on this matrix, sparse masks are constructed by selecting the most important blocks or clusters. 
Among these methods, SVG2~\cite{yang2025sparse} groups tokens by feature similarity via $k$-means clustering and employs cluster centroids to estimate attention scores for cluster selection, achieving state-of-the-art performance.

While this clustering strategy is efficient, its optimization objective is not fully aligned with minimizing the importance estimation error (see Section~\ref{sec:motivation_interaction} for details). A more principled approach is to cluster tokens based on their query-key dot-product interactions. 
However, explicitly clustering tokens by their full query-key dot-product interactions is prohibitively expensive for long video sequences: constructing the full query-key dot-product matrix already incurs quadratic complexity in sequence length, not to mention that each interaction vector itself also has sequence-length dimensionality. To tackle this issue, we reformulate the clustering objective derived from query-key interactions as a quadratic form in the original query/key space, avoiding explicit construction of the full matrix. This yields a more efficient clustering scheme for more accurate importance estimation.

Upon the estimated importance, sparse masks are determined by pre-defined rules. The fixed top-\(p\) rule~\cite{zhu2026tactic, NEURIPS2025_ad2f5711} is widely used by selecting the highest-scoring query-key regions until the cumulative importance reaches \(p\). While simple and effective, this rule yields a similar sparsity across the denoising process, overlooking the fact that tolerance to sparse approximation errors varies across timesteps. 
Since signal-to-noise ratio (SNR) characterizes the denoising state~\cite{ho2020denoising, song2021scorebased, lipman2022flow} by reflecting the relative strength of noise and signal at different timesteps, it provides a natural basis for adaptive sparsity control. Based on this, we propose an SNR-guided sparsity schedule, in which the top-\(p\) value is computed offline from a smooth function of the logarithmic SNR, effectively driving sparse masks to match various denoising timesteps.

When deploying these optimization methods on hardware, \textbf{execution efficiency} is critical. In existing cluster-based sparse attention, mapping clusters to block-based GPU kernels often produces underfilled tail blocks due to irregular cluster sizes~\cite{dao2022flashattention, dao2023flashattention, shah2024flashattention, zadouri2026flashattention4algorithmkernelpipelining}, increasing computational overhead and reducing hardware utilization. This problem is particularly severe on the query side, where the same padding overhead may be incurred repeatedly across multiple selected key clusters. To alleviate this, we introduce a tail-merging strategy that reduces padding overhead and improves utilization during sparse attention execution.

Building on these designs, we propose iCATS (Fast Video Generation via Interaction-Aware Sparse Attention and Timestep-Adaptive Sparsity), a training-free framework that performs optimization throughout the sparse attention pipeline. The main contributions of this work are as follows:

\begin{itemize}
    \item We propose a training-free method, called iCATS, for DiT acceleration. iCATS clusters tokens according to query-key interactions via a quadratic-form objective defined in the original query/key space, avoiding complicated full query-key dot-product computation while improving the accuracy of importance estimation.

    \item For sparse mask construction, iCATS introduces an SNR-guided sparsity schedule that dynamically adjusts the top-$p$ value across denoising timesteps, thereby allocating the computation budget to the selected informative query-key interactions.

    \item Moreover, when deployed on hardware, iCATS develops a tail-merging strategy that reduces padding overhead from irregular query cluster sizes, improving GPU kernel utilization during sparse attention execution.
    
    \item Extensive experiments on representative DiT-based video generation models show that iCATS achieves state-of-the-art performance among training-free sparse attention methods, delivering an advanced trade-off between generation quality and inference efficiency.
\end{itemize}

\section{Related Work}
\label{sec:Related Work}


\subsection{Training-based Sparse Attention}
Training-based sparse attention methods accelerate DiT-based video generation by learning model-adaptive sparse computation patterns. DSV~\cite{tan2025dsv} trains separate low-rank predictors to identify critical key-value pairs and achieves acceleration using specialized kernels and sparsity-aware parallelism. However, the decoupled predictor design limits end-to-end optimization. VSA~\cite{NEURIPS2025_dfc310e8} addresses this issue by introducing an end-to-end trainable sparse attention design: it first performs coarse cube-level attention on pooled representations to locate important regions, and then computes token-level attention within the selected cube blocks, enabling joint optimization of selection and computation. VMoBA~\cite{wu2026vmoba} likewise adopts block-level selection, but with a different motivation: it exploits the spatio-temporal locality of videos through recurrent 1D-2D-3D block partitioning, global block selection, and threshold-based selection.

The above methods all directly discard unselected tokens, whereas SLA~\cite{zhang2026sla} revisits this design and instead partitions blocks into three computational levels: sparse attention for critical blocks, linear attention for marginal blocks, and direct skipping for negligible ones. Building on this idea, SLA2~\cite{zhang2026sla2} replaces heuristic splitting with a learnable router and further introduces ratio-based branch fusion and quantization-aware low-bit sparse attention.

Despite their different designs, these methods all rely on additional model-specific training or fine-tuning, which limits their flexibility when the target model changes.

\subsection{Training-free Sparse Attention}

Training-free sparse attention improves inference efficiency without modifying model parameters, which is more flexible for deployment. One line of work uses predefined sparse patterns derived from spatio-temporal priors. SVG~\cite{xi2025sparse} profiles per-head spatial and temporal patterns online and applies the resulting masks through FlexAttention~\cite{dong2024flexattentionprogrammingmodel}. RainFusion~\cite{chen2025rainfusion} and Sparse-vDiT~\cite{chen2025sparsevdit} extend this line with richer predefined patterns, while STA~\cite{zhang2025fast} replaces full attention with sliding tile attention over local 3D neighborhoods. Although these methods are fast, their fixed or hand-designed sparse patterns limit adaptivity to diverse video content.

Another line of work dynamically constructs sparse masks based on token features. SpargeAttn~\cite{zhang2025spargeattention} identifies important blocks through a two-stage online filtering scheme. XAttention~\cite{xu2025xattention} estimates block importance using strided antidiagonal sums followed by cumulative thresholding. SVG2~\cite{yang2025sparse} further improves dynamic sparse attention by first applying $k$-means clustering to group tokens with similar features, then using cluster centroids to estimate attention importance for group selection, and finally reordering the selected groups into contiguous layouts for more efficient sparse attention execution. Compared with predefined-pattern methods, these dynamic approaches are more content-adaptive. However, their token grouping operates independently in the query and key feature spaces; while simple, it sacrifices accuracy.

Following importance estimation, most existing methods~\cite{xi2025sparse, yang2025sparse} apply a fixed sparsity level across denoising timesteps to construct sparse masks. This produces similar sparsity patterns across steps, even though different denoising timesteps exhibit distinct characteristics during the generation process. These observations motivate a more accurate and efficient sparse attention framework that optimizes the entire pipeline.



\section{Motivation}
\label{sec:Motivation}

\subsection{Independent Clustering Mismatches Q-K Interaction Objectives}
\label{sec:motivation_interaction}

As aforementioned, existing sparse attention methods~\cite{yang2025sparse} typically adopt a clustering-based approach for importance estimation: clustering centroids are used to construct a smaller query-key matrix that replaces the original full matrix. Therefore, the clustering objective is to minimize the error between the original and the smaller query-key matrices.

To further explain, we fix the query matrix $Q$ and analyze key clustering as an illustrative example. Let $G=\{B_t\}_{t=1}^{P}$ denote a partition of key tokens, where each key token $k_j \in \mathbb{R}^{1\times d}$ is a row vector and $\mu_t \in \mathbb{R}^{1\times d}$ is the centroid of cluster $B_t$. If each key token is replaced by its cluster centroid when forming the smaller query-key matrix, the optimization objective is then to minimize the resulting error in the query-key dot-product interactions, i.e., minimizing $E(G)=\sum_{t=1}^{P}\sum_{j\in B_t}\|Q(k_j^\top-\mu_t^\top)\|_2^2$. By contrast, previous key-space clustering methods like $k$-means assume that $Q^\top Q \propto I$, i.e., the query matrix treats all directions in the key space isotropically, and simplify $E(G)$ to ${\hat E}(G)=\sum_{t=1}^{P}\sum_{j\in B_t}\|k_j-\mu_t\|_2^2$. Apparently, this $\hat E(G)$ optimization is not aligned with the true objective $E(G)$ of minimizing the query-key interaction error, thereby reducing estimation accuracy.

Instead, this paper proposes to directly optimize $E(G)$.
However, this is computationally expensive for long video sequences, as it requires constructing the full query-key dot-product matrix and representing each token with a sequence-length interaction vector, which increases inference time. Therefore, we develop an efficient method, termed \textbf{Interaction-aware Clustering}, which minimizes $E(G)$ using a low-complexity scheme without explicitly constructing the full matrix.

\subsection{Fixed Sparsity Overlooks Denoising Dynamics}
\label{sec:motivation_Sparsity}

In flow-based video generation~\cite{lipman2022flow}, samples are synthesized through an iterative denoising process. Let a latent representation \(x_0 \in \mathbb{R}^{T \times H \times W \times C}\) be mapped from a clean video \(X\), where \(T\), \(H\), \(W\), and \(C\) denote the numbers of frames, height, width, and channels. At denoising timestep \(t \in \{0,\dots,N\}\), the latent is $x_t = a_t x_0 + b_t \epsilon$,
where \(\epsilon \sim \mathcal{N}(0,I)\), \(N\) is the total number of solver steps, and \(a_t\) and \(b_t\) are determined by the timestep schedule. As \(t\) increases, \(x_t\) transitions from signal-dominated to noise-dominated, while sampling follows the reverse denoising process back to \(x_0\).

Existing methods typically use a fixed top-$p$ rule to construct sparse masks across denoising timesteps. Although the exact number of selected blocks or clusters may vary slightly due to score estimation, the overall sparsity remains broadly similar throughout the denoising trajectory (see our analysis in appendix~\ref{app:density}). 
While simple, this implicitly assumes that approximation errors between dense and sparse computations contribute uniformly across timesteps. However, denoising is a recursive process in which the state at each timestep depends on the previous update. Consequently, errors introduced at one timestep accumulate and propagate along the denoising trajectory. As a result, errors at different timesteps have varying impacts on the final output.


Therefore, under a fixed computational budget, the key challenge is not only how much attention computation can be reduced but also how to allocate this budget appropriately across timesteps to best preserve quality. This motivates our \textbf{SNR-guided Sparsity} that adjusts sparse mask construction according to the characteristics of each denoising timestep. As the SNR directly reflects denoising progress and signal dominance, it serves as an ideal indicator for such adaptation.

\subsection{Varying Cluster Sizes Incur Tail Overhead}
\label{sec:motivation_Tail}





\begin{wrapfigure}{r}{0.45\linewidth}
  \vspace{-3mm}
  \centering
  \subfloat[block-based]{%
    \includegraphics[width=0.5\linewidth]{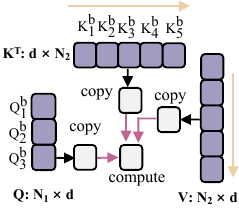}%
    \label{fig:block_based}%
  }\hfill
  \subfloat[cluster-based]{%
    \includegraphics[width=0.5\linewidth]{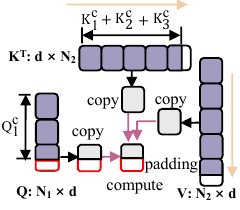}%
    \label{fig:cluster_based}%
  }
  \caption{Visualization of (a) block-based and (b) cluster-based sparse attention.}
  \label{fig:block_vs_cluster}
\end{wrapfigure}

Unlike block-based sparse attention, which naturally matches fixed-size block kernels (Figure~\ref{fig:block_based}), cluster-based sparse attention selects computation at the cluster level (Figure~\ref{fig:cluster_based}). This easily produces irregular cluster sizes that do not align with the fixed execution granularity of sparse kernels.

More specifically, in cluster-based sparse attention with block-based kernels, tokens from selected key clusters are gathered into a contiguous sequence for each query cluster, and attention is computed block by block. The last block is often partially filled, yet the kernel still executes full-block computation, introducing overhead on padded positions. This inefficiency is especially severe on the query side, where padding in the last query block repeats across all key/value blocks, causing overhead to scale with the selected key length. Therefore, reducing tail execution overhead is crucial for improving hardware efficiency, motivating our \textbf{Tail-merging Strategy}.

\section{Method}
\label{sec:Method}

\begin{figure}[t]
    \centering
    \includegraphics[width=\linewidth]{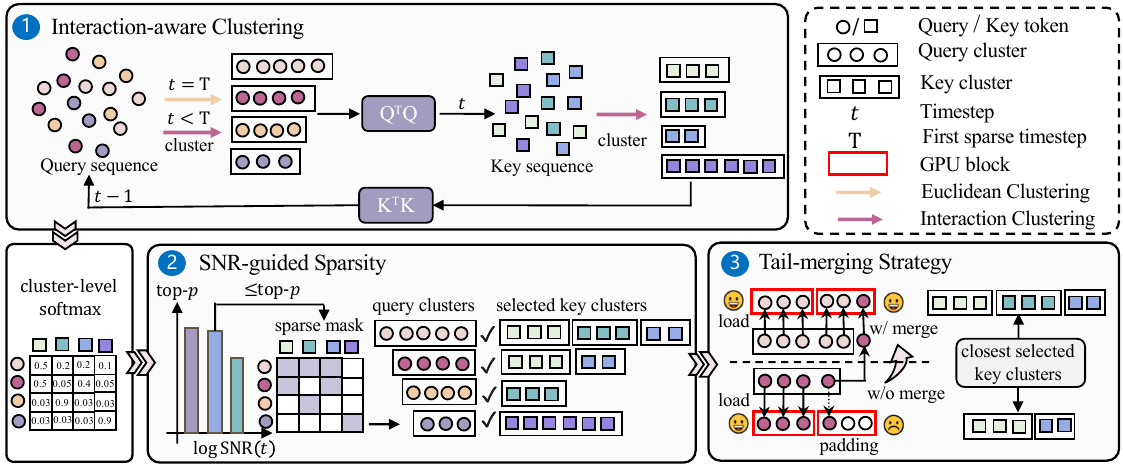}
    \caption{Overview of iCATS pipeline. For each attention layer, query and key tokens are first grouped by \textbf{\ding{172} Interaction-aware Clustering}. Then, the corresponding query-key cluster centroids are used to approximate softmax scores at the cluster level. Based on these approximated scores, \textbf{\ding{173} SNR-guided Sparsity} determines the top-$p$ value at each denoising timestep and selects the corresponding attention computation region. Finally, \textbf{\ding{174} Tail-merging Strategy} is applied before sparse attention kernel computation by reassigning tail tokens of non-block-aligned query clusters, thereby reducing padding overhead caused by irregular cluster sizes. In this figure, interaction clustering denotes interaction-aware query-key clustering, while Euclidean clustering is used to initialize query clusters solely at the first sparse timestep.}
    \label{fig:pipeline}
\end{figure}

This section details our proposed iCATS. As shown in Figure~\ref{fig:pipeline}, iCATS consists of three components: interaction-aware clustering for accurate importance estimation (Section~\ref{sec:method_clustering}), SNR-guided sparsity for sparse mask construction (Section~\ref{sec:method_timestep}), and tail merging for sparse attention execution (Section~\ref{sec:method_tail}).

\subsection{Interaction-Aware Clustering}
\label{sec:method_clustering}

Recalling Section~\ref{sec:motivation_interaction}, iCATS performs interaction-aware clustering by grouping tokens using their query-key dot-product interactions. We represent all token features as row vectors. Let \(Q=[q_1^\top,q_2^\top,\dots,q_n^\top]^\top \in \mathbb{R}^{n\times d}\) (\(d \ll n\)) denote the query matrix, where \(q_i \in \mathbb{R}^{1\times d}\) is the \(i\)-th query token. Here, \(n\) is the sequence length and \(d\) is the token dimension, e.g., \(n=75,600\) and \(d=128\) in Wan2.1-T2V-14B.  Let \(B_t\) denote the \(t\)-th key cluster and \(\mu_t \in \mathbb{R}^{1\times d}\) its centroid, where \(t\in\{1,\dots,P\}\) with \(P\) being the total number of key clusters. We define the key clustering objective as:
\begin{equation}
\begin{aligned}
\mathcal{L}
&=\sum_{t=1}^{P}\sum_{j\in B_t}
\left\|Qk_j^\top-Q\mu_t^\top\right\|_2^2
=\sum_{t=1}^{P}\sum_{j\in B_t}\sum_{i=1}^{n}
\left(q_i k_j^\top - q_i \mu_t^\top\right)^2 \\
&=\sum_{t=1}^{P}\sum_{j\in B_t}
(k_j-\mu_t)\left(\sum_{i=1}^{n} q_i^\top q_i\right)(k_j^\top-\mu_t^\top)
=\sum_{t=1}^{P}\sum_{j\in B_t}
(k_j-\mu_t)\,Q^\top Q\,(k_j^\top-\mu_t^\top).
\end{aligned}
\label{eq:key_cluster_obj}
\end{equation}
As seen, clustering using query-key interactions requires computing, for each key, its dot products with all queries to form an \(n\)-dimensional interaction vector, followed by clustering in this space. Equation~\eqref{eq:key_cluster_obj} finally reformulates this objective as a quadratic form in the original key feature space of dimension \(d\), where \(d \ll n\), avoiding explicit construction of the full query-key dot-product matrix. Query clustering is defined symmetrically, yielding a corresponding quadratic form induced by \(K^\top K\). Although \(Q^\top Q\) and \(K^\top K\) are only of size \(d\times d\), computing them still requires aggregating statistics over all tokens. To reduce this overhead, we approximate them using cluster centroids and sizes. Specifically, for query clusters \(\{A_r\}_{r=1}^{O}\), where \(O\) is the number of query clusters, with centroid \(\mu_r\) and size \(|A_r|\), and key clusters \(\{B_t\}_{t=1}^{P}\) defined similarly with size \(|B_t|\), we define \(M_Q=\sum_{r=1}^{O}|A_r|\,\mu_r^\top \mu_r\) and \(M_K=\sum_{t=1}^{P}|B_t|\,\mu_t^\top \mu_t\).
Approximating \(Q^\top Q\) with \(M_Q\) and \(K^\top K\) with \(M_K\), and noting that \(M_Q\) and \(M_K\) are symmetric matrices, the quadratic-form distances for clustering become:
\begin{equation}
\begin{aligned}
(k_j-\mu_t)M_Q(k_j^\top-\mu_t^\top)
&= k_j M_Q k_j^\top + \mu_t M_Q \mu_t^\top - 2\,k_j M_Q \mu_t^\top,\\
(q_i-\mu_r)M_K(q_i^\top-\mu_r^\top)
&= q_i M_K q_i^\top + \mu_r M_K \mu_r^\top - 2\,q_i M_K \mu_r^\top.
\end{aligned}
\label{eq:quad_dist}
\end{equation}
Built on Flash-Kmeans~\cite{yang2026flashkmeansfastmemoryefficientexact}, we precompute the token-side terms (e.g., \(k_j M_Q\), \(k_j M_Q k_j^\top\), \(q_i M_K\), and \(q_i M_K q_i^\top\)) only once before the clustering iterations begin, as token features remain fixed. The centroid-side terms are recomputed from the current centroids at each iteration.

Based on this design, we perform interaction-aware clustering alternately. At the first sparse timestep, query tokens are clustered in the original feature space, and the resulting query centroids and cluster sizes are used to construct \(M_Q\). Key tokens are then clustered using the quadratic form induced by \(M_Q\), and the resulting key centroids and cluster sizes are used to construct \(M_K\). At later timesteps, we use the previous \(M_K\) to guide query clustering, leveraging the similarity of quadratic structures between adjacent timesteps. The updated query clusters define the current \(M_Q\), which in turn induces the quadratic form for key clustering. The resulting key clusters form a new \(M_K\) for the next timestep.

\subsection{SNR-Guided Sparsity}
\label{sec:method_timestep}
Motivated by the observation in Section~\ref{sec:motivation_Sparsity}, iCATS devises the SNR-guided sparsity schedule, which dynamically adjusts the sparsity of sparse masks across timesteps. Specifically, we employ the timestep-wise SNR to characterize the denoising state: a low SNR typically corresponds to noise-dominated latents at earlier denoising stages, where approximation errors may affect more subsequent updates, while a high SNR indicates the emergence of cleaner signal components and more stable semantic structures. This makes SNR a practical indicator for adapting top-$p$ across timesteps. Formally, the timestep-wise SNR is 
defined as $\mathrm{SNR}(t)=\left({a_t}/{b_t}\right)^2$.

\begin{figure}[t]
    \centering
    \subfloat[SNR]{%
        \includegraphics[width=0.28\linewidth]{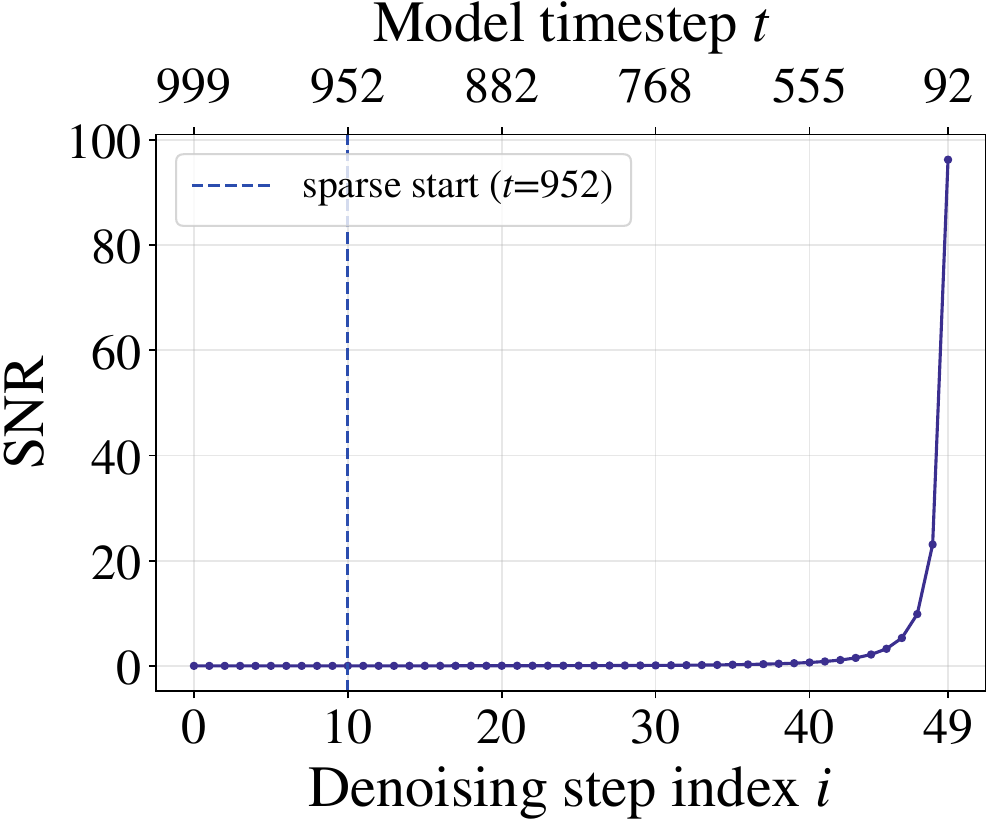}%
        \label{subfig:snr_curve}%
    }%
    \hspace{0.02\linewidth}
    \subfloat[$\mathrm{logSNR}(t)$]{%
        \includegraphics[width=0.28\linewidth]{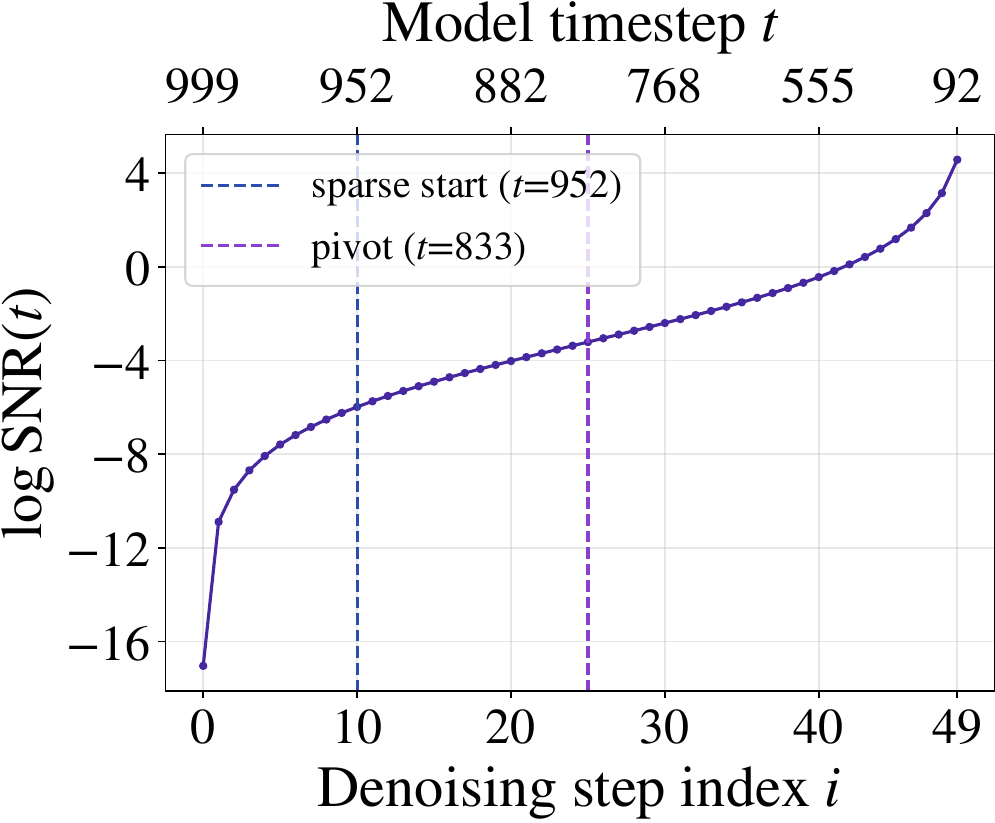}%
        \label{subfig:log_snr_curve}%
    }%
    \hspace{0.02\linewidth}
    \subfloat[Top-$p$ from $\mathrm{logSNR}(t)$]{%
        \includegraphics[width=0.28\linewidth]{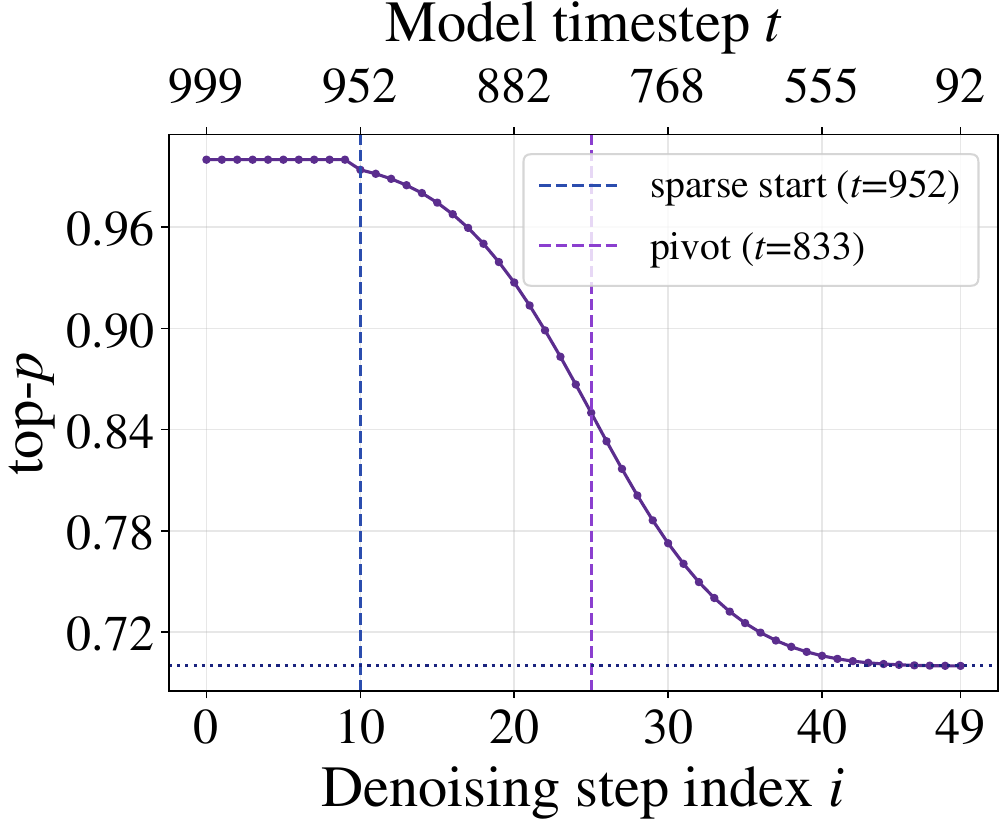}%
        \label{subfig:top_p_curve}%
    }

    \caption{Visualization of proposed SNR-guided sparsity schedule on Wan2.1-T2V-14B.
    (a) Raw SNR over denoising timesteps.
    (b) \(\mathrm{logSNR}(t)\), where the pivot marks the center of the flattest region.
    (c) The resulting top-$p$ schedule generated by an inverted sigmoid.}
    \label{fig:snr_schedule}
    \vspace{-10pt}
\end{figure}

Figure~\ref{subfig:snr_curve} shows that the raw SNR spans a wide range across timesteps, making early- and late-stage differences hard to distinguish. We therefore operate in the logarithmic domain and define \(\ell_t=\log \mathrm{SNR}(t)\) (see Figure~\ref{subfig:log_snr_curve}). This converts ratio changes in SNR into additive differences, making variations across timesteps easier to model and analyze. 

As analyzed in Figure~\ref{subfig:log_snr_curve}, the $\mathrm{logSNR}(t)$ curve exhibits a clear stage-wise pattern: the early and late stages change more sharply, while the middle stage is comparatively smooth. Based on this observation, we use an inverted $\mathrm{sigmoid}$ function to generate timestep-adaptive top-$p$ values for sparse mask construction. The inverted $\mathrm{sigmoid}$ provides a simple schedule shape that aligns well with the stage-wise pattern of the $\mathrm{logSNR}(t)$ curve: its two saturated ends help maintain relatively stable computation in the early and late stages, while its transition region allows top-$p$ to decrease gradually in the smoother middle stage. Overall, we keep dense attention before the first sparse timestep \(t_{\mathrm{sparse}}\), and apply sparse attention for \(t \le t_{\mathrm{sparse}}\) with
\begin{equation}
\mathrm{top}\text{-}p_t =
p_{\min} + (1-p_{\min})\left(1-\frac{1}{1+\exp(-k x_t)}\right),
\qquad t \le t_{\mathrm{sparse}},
\label{eq:sigmoid_schedule}
\end{equation}

\vspace{-10pt}
where \(p_{\min}\) denotes the minimum top-$p$ value, and \(x_t = \ell_t - \ell_{t_{\mathrm{pivot}}}\) is defined by the $\mathrm{logSNR}(t)$ difference between timestep \(t\) and a pivot timestep \(t_{\mathrm{pivot}}\). The pivot is chosen automatically as the timestep immediately following the minimum difference between consecutive $\mathrm{logSNR}(t)$ values, which typically lies near the smoothest denoising stage.

We choose $\tau \in (0,1)$ to be very close to 1 in practice (e.g., $0.99$ or $0.999$), which controls how close the inverted $\mathrm{sigmoid}$  is to its upper bound at the first sparse timestep and thus determines the sigmoid slope as $k = \frac{\log\frac{\tau}{1-\tau}}{\left|\,\ell_{t_{\mathrm{sparse}}}-\ell_{t_{\mathrm{pivot}}}\,\right|}$. Finally, rather than introducing an additional parameter, we set the minimum retained ratio of key clusters by linearly scaling top-$p$ as $\text{min-}p_t = 0.1 \cdot \frac{\text{top-}p_t}{\text{top-}p_{t_{\mathrm{pivot}}}}$.



As illustrated in Figure~\ref{subfig:top_p_curve}, the final top-$p$ schedule resulting from $\mathrm{logSNR}(t)$ keeps the sparse masking budget more stable in the early and late stages while reducing top-$p$ more smoothly in the middle stage, which better aligns with the stage-wise characteristics of denoising. Since the SNR sequence is fully determined by the predefined noise schedule, the top-$p$ values can be computed offline before video generation, largely saving runtime overhead during inference. To further illustrate this behavior, we visualize the computational density of iCATS across timesteps in Appendix~\ref{app:density} (Figure~\ref{suppfig:density}), where density denotes the  proportion of retained computations.

\subsection{Tail-Merging Strategy}
\label{sec:method_tail}

Motivated by the tail execution issue analyzed in Section~\ref{sec:motivation_Tail}, iCATS develops a tail-merging strategy before the kernel computation of sparse attention to reduce the extra overhead caused by irregular cluster sizes. The core idea is to restructure query clusters in a hardware-friendly manner while preserving the original sparse attention pattern as much as possible, thereby reducing both scheduling and padding overhead. To this end, we first exploit merging opportunities that do not alter the original sparse pattern, and then further reduce the remaining tail waste.

Specifically, we merge query clusters within the same attention head that attend to the same set of key clusters. Because these clusters share the same attended key-cluster set, this operation preserves the original sparse attention pattern without introducing additional error, while reducing the number of query clusters and subsequent scheduling overhead.

We then handle query clusters whose sizes are not divisible by the block execution size, since their final execution blocks are only partially occupied while still fully computed, leading to tail overhead. Our goal is to reassign only tokens in these tail blocks to improve execution-block utilization while minimizing perturbation to the sparse attention pattern. To this end, we need to find suitable target clusters for these tail tokens. Specifically, we measure the compatibility between query clusters based on the similarity of their key-cluster masks. Each mask is represented as a packed binary vector, from which pairwise Hamming distances between clusters are computed. A smaller Hamming distance indicates that the key-cluster masks of two query clusters are more similar; when the merged mask is updated by union, this usually leads to fewer newly introduced key clusters, making them better candidates for merging. Under size and alignment constraints, we iteratively identify compatible cluster pairs and reassign the tail-block tokens of the source cluster to the target cluster. After merging, empty clusters are removed, and the remaining clusters are compactly reindexed. In implementation, we perform two rounds of tail merging. Detailed pseudocode is presented in Appendix~\ref{app:algo_over}.



\section{Experiments and Analysis}
\label{sec:Experiment}

\subsection{Setup}

\textbf{Experimental setup}. 
We evaluate iCATS on three mainstream video generation DiT models, including HunyuanVideo-T2V-13B, Wan2.1-T2V-14B, and Wan2.1-I2V-14B, for 5-second 720p generation. Following prior works~\cite{xi2025sparse, yang2025sparse}, we use the Penguin Benchmark with VBench~\cite{huang2023vbench, zheng2025vbench2} prompt optimization for text-to-video evaluation. For image-to-video evaluation, we use VBench prompt-image pairs and resize the input images to 16:9 to match the 720p setting. Ablations are conducted on Wan2.1-T2V-14B using 138 randomly sampled text prompts. More setting details can be found in Appendix~\ref{app:settings}.

\textbf{Evaluation metrics}. 
We measure the similarity between videos generated by the accelerated DiT models and the original DiT models using PSNR, SSIM~\cite{wang2004image}, and LPIPS~\cite{zhang2018unreasonable} and evaluate video quality with VBench~\cite{huang2023vbench}. The main paper reports the average VBench score, while the detailed breakdown over subject consistency, background consistency, motion smoothness, aesthetic quality, and imaging quality is reported in Appendix~\ref{app:vbench}. To assess practical efficiency, we report the speedup over the original dense attention. We also report sparsity, defined as the theoretical computation reduction ratio of sparse attention methods relative to dense attention.

\textbf{Comparison setup}. 
We compare iCATS against two state-of-the-art sparse attention methods, SVG~\cite{xi2025sparse} and SVG2~\cite{yang2025sparse}, using their official open-source implementations and settings for fair comparisons. Moreover, we evaluate two iCATS variants, Base and Turbo, whose hyperparameters are chosen to match the speed or sparsity of prior methods. Unless otherwise stated, experiments are conducted on eight NVIDIA H100 GPUs.

\subsection{Quality \& Efficiency Evaluation}

Table~\ref{tab:main_results} shows that iCATS-Base and iCATS-Turbo consistently achieve a better quality-efficiency trade-off than SVG and SVG2 on both HunyuanVideo-T2V-13B and Wan2.1-T2V-14B. At comparable speedup, iCATS-Base achieves better PSNR, SSIM, LPIPS, and VBench than SVG2, indicating stronger quality preservation. Meanwhile, iCATS-Turbo provides higher acceleration at similar sparsity while maintaining better reconstruction quality and competitive VBench scores. The same trend also holds on Wan2.1-I2V-14B, where iCATS-Base improves all reconstruction metrics and VBench over SVG2 at comparable speed, while iCATS-Turbo achieves the fastest inference among sparse methods and still maintains better reconstruction quality with matched VBench. 
Detailed VBench results can be found in Appendix~\ref{app:vbench}. Visualization comparisons are shown in Figure~\ref{fig:visual} and Appendix~\ref{app:visual}. Results of cross-model and cross-task hyperparameter transfer without retuning are provided in Appendix~\ref{app:hyperparameter_transfer}.

\begin{table}[h]
\centering
\caption{Comparison with existing works on various benchmark models}
\label{tab:main_results}
\small
\setlength{\tabcolsep}{4pt}
\begin{tabularx}{\textwidth}{>{\raggedright\arraybackslash}Xcccccc}
\toprule
\textbf{Method} & \textbf{PSNR}$\uparrow$ & \textbf{SSIM}$\uparrow$ & \textbf{LPIPS}$\downarrow$ & \textbf{VBench}$\uparrow$ & \textbf{Time (s)}$\downarrow$ & \textbf{Sparsity} \\
\midrule
HunyuanVideo-T2V-13B (Dense)  & --     & --     & --     & 0.844 & 1730               & --    \\
\midrule
SVG                           & 20.362 & 0.725  & 0.235  & 0.843 & 874 (1.98$\times$)  & 75.8\% \\
SVG2                          & 26.247 & 0.869  & 0.097  & 0.842 & 853 (2.03$\times$)  & 71.5\% \\
\rowcolor{blue!10}
iCATS-Base                    & \textbf{31.017} & \textbf{0.926} & \textbf{0.050} & \textbf{0.844} & 854 (2.03$\times$)  & 66.1\% \\
\rowcolor{blue!20}
iCATS-Turbo                   & 30.466 & 0.922  & 0.054  & \textbf{0.844} & 827 (2.09$\times$)  & 70.3\% \\
\midrule
Wan2.1-T2V-14B (Dense)        & --     & --     & --     & 0.839 & 1889               & --    \\
\midrule
SVG                           & 21.497 & 0.741  & 0.197  & 0.836 & 1220 (1.55$\times$) & 65.7\% \\
SVG2                          & 25.606 & 0.853  & 0.098  & 0.837 & 1226 (1.54$\times$) & 68.6\% \\
\rowcolor{blue!10}
iCATS-Base                    & \textbf{29.301} & \textbf{0.908} & \textbf{0.058} & \textbf{0.838} & 1222 (1.55$\times$) & 61.7\% \\
\rowcolor{blue!20}
iCATS-Turbo                   & 28.172 & 0.895  & 0.068  & 0.837 & 1176 (1.61$\times$) & 69.0\% \\
\midrule
Wan2.1-I2V-14B (Dense)        & --     & --     & --     & 0.847 & 1519               & --    \\
\midrule
SVG                           & 26.533 & 0.847  & 0.101  & 0.844 & 1088 (1.40$\times$) & 65.7\% \\
SVG2                          & 29.082 & 0.890  & 0.067  & 0.845 & 1081 (1.41$\times$) & 69.9\% \\
\rowcolor{blue!10}
iCATS-Base                    & \textbf{30.606} & \textbf{0.911} & \textbf{0.056} & \textbf{0.846} & 1084 (1.41$\times$) & 63.2\% \\
\rowcolor{blue!20}
iCATS-Turbo                   & 30.097 & 0.905  & 0.060  & \textbf{0.846} & 1057 (1.44$\times$) & 68.9\% \\
\bottomrule
\end{tabularx}
\end{table}

To understand the quality gains from interaction-aware clustering, we compare both clustering schemes with exact dense attention using two metrics. (i) Token-level top-$p$ selection overlap: for each query token $q$, we compute its exact dense attention weights (full softmax over all keys) and apply top-$p$ ($p=0.9$) to obtain the keys selected for $q$; the method computes keys in the key clusters selected for the query cluster containing $q$; overlap is the fraction of dense-selected keys covered by the method, averaged over all query tokens. (ii) Attention output MSE between dense and masked attention outputs (attention weights multiplied by $V$), with masked weights zeroed and the remaining weights renormalized. The diagnostics are conducted on Wan2.1-T2V-14B with 20 randomly sampled prompts, covering 5 timesteps ($t = 952, 882, 768, 555, 92$), 5 layers (1, 10, 20, 30, 39), and 4 heads (0, 10, 20, 30) per layer. Both schemes share the same top-$p$ ($0.9$), min-$p$ ($0.1$), cluster configuration (400 query / 1000 key clusters), clustering iterations (50 at the first sparse timestep, then 2), warmup, and kernel implementation; only the clustering scheme differs. As shown in Table~\ref{tab:cluster_fidelity}, interaction-aware clustering achieves higher selection overlap (+1.7 points) and a 36\% lower output MSE, with consistent gains at every sampled timestep, layer, and head (Appendix~\ref{app:fidelity_breakdown}). These results indicate that interaction-aware clustering better matches dense attention in both region selection and attention outputs. On Wan2.1-T2V-14B over 20 randomly sampled prompts, replacing $M_Q/M_K$ with exact $Q^\top Q/K^\top K$ under the same SNR-guided sparsity schedule yields the same reported token-level overlap (0.912 for both), while increasing denoising time from 1219 s to 1257 s.

\begin{table}[h]
\centering
\caption{Attention fidelity of different clustering schemes on Wan2.1-T2V-14B.}
\label{tab:cluster_fidelity}
\small
\renewcommand{\arraystretch}{1.18}
\setlength{\tabcolsep}{6pt}
\begin{tabular}{lcc}
\toprule
\textbf{Method} & \textbf{Top-$p$ overlap}$\uparrow$ & \textbf{Output MSE ($\times 10^{-4}$)}$\downarrow$ \\
\midrule
Independent Euclidean clustering & 0.896 & 8.99 \\
\midrule
\rowcolor{blue!10}
Interaction-aware clustering (ours) & \textbf{0.913} & \textbf{5.75} \\
\bottomrule
\end{tabular}
\end{table}

\begin{wrapfigure}{r}{0.34\linewidth}
  \vspace{-1em}
  \centering
  \includegraphics[width=\linewidth]{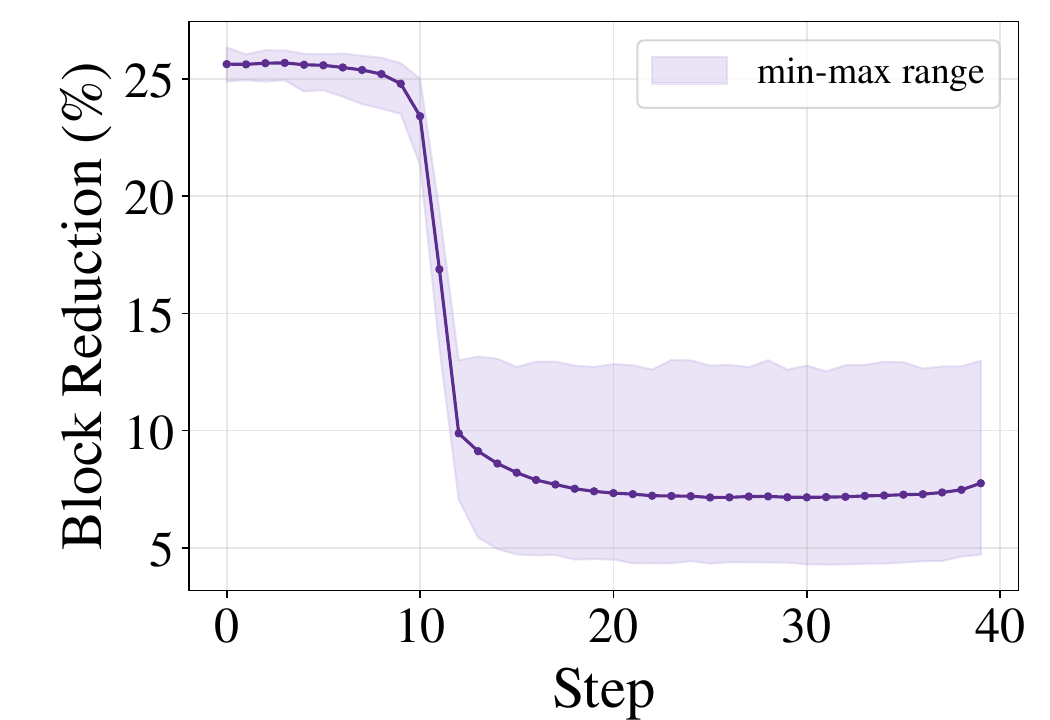}
  \vspace{-3.5mm}
  \caption{Reduction ratio of execution blocks on Wan2.1-T2V-14B.}
  \label{fig:block_reduction}
\end{wrapfigure}

To quantify the effect of tail merging, we measure the reduction in execution blocks before and after merging at each sparse attention timestep, where execution blocks are defined following FlashInfer~\cite{ye2025flashinfer}. Figure~\ref{fig:block_reduction} shows the average reduction ratio of iCATS, with the shaded region indicating the min-max range across layers. The reduction is largest at early timesteps and gradually decreases during denoising. 
The merging procedure is efficiently implemented in Triton~\cite{tillet2019triton}.  On Wan2.1-T2V-14B running on an NVIDIA H100 GPU, each invocation of the merging function takes $\sim$3 ms, while merging itself takes $\sim$10 s per generation, which is negligible relative to the total inference time of 1222 s. More details are provided in Appendix~\ref{app:merge_impl}. A detailed breakdown of the denoising runtime is provided in Appendix~\ref{app:runtime}.

\subsection{Cross-Hardware Validation}
To evaluate the hardware generalization of iCATS, we further deploy iCATS-Turbo on Wan2.1-T2V-14B on the T-Head Zhenwu M890P\footnote{Technical documentation: \url{https://developer.t-head.cn/docs_center/doc_detail/index.html?projectId=48}}, an AI accelerator developed by Alibaba T-Head, with sparse kernel operators adapted to this platform. As shown in Table~\ref{tab:wan21_t2v_14b_ZHENWU_results}, iCATS achieves performance comparable to that on H100 GPUs, demonstrating strong adaptability and generalization across hardware platforms.

\vspace{-3mm}
\begin{table}[h]
\centering
\caption{Comparison of different methods on Wan2.1-T2V-14B using the Zhenwu M890P.}
\label{tab:wan21_t2v_14b_ZHENWU_results}
\small
\renewcommand{\arraystretch}{1.18}
\setlength{\tabcolsep}{4pt}
\begin{tabularx}{\textwidth}{>{\raggedright\arraybackslash}Xcccccc}
\toprule
\textbf{Method} & \textbf{PSNR}$\uparrow$ & \textbf{SSIM}$\uparrow$ & \textbf{LPIPS}$\downarrow$ & \textbf{Time (s)}$\downarrow$ & \textbf{Speedup}$\uparrow$ & \textbf{Sparsity} \\
\midrule
Wan2.1-T2V-14B (Dense) & -- & -- & -- & 2828 & -- & -- \\
\midrule
SVG                    & 21.847 & 0.774 & 0.186 & 2208 & 1.28$\times$ & 65.7\% \\
SVG2                   & 26.321 & 0.870 & 0.088 & 2214 & 1.28$\times$ & 68.9\% \\
\rowcolor{blue!20}
iCATS-Turbo            & \textbf{28.849} & \textbf{0.903} & \textbf{0.061} & 2073 & 1.36$\times$ & 68.9\% \\
\bottomrule
\end{tabularx}
\end{table}

\subsection{Human Evaluation}
\label{sec:human_evaluation}

We blindly compare iCATS-Base with SVG2 on Wan2.1-T2V-14B using 20 VBench-optimized Penguin Benchmark prompts covering complex scenes, highly dynamic human actions, and simple static scenes. Eleven video-domain professionals with normal vision view anonymized, randomly ordered A/B pairs side by side on 2K screens, selecting the perceptually better video or ``Similar Quality,'' yielding 220 votes. Table~\ref{tab:human_evaluation} shows 54.1\% preference for iCATS-Base, 18.6\% for SVG2, and 27.3\% similar quality, supporting iCATS-Base’s perceptual advantage on these samples beyond automatic metrics.
\vspace{-3mm}
\begin{table}[htbp]
\centering
\caption{Human preference on Wan2.1-T2V-14B.}
\label{tab:human_evaluation}
\small
\renewcommand{\arraystretch}{1.15}
\setlength{\tabcolsep}{12pt}
\begin{tabular}{lcc}
\toprule
\textbf{Preference} & \textbf{Votes} & \textbf{Percentage} \\
\midrule
SVG2            & 41  & 18.6\% \\
\rowcolor{blue!20}
\textbf{iCATS-Base}      & \textbf{119} & \textbf{54.1\%} \\
Similar Quality & 60  & 27.3\% \\
\bottomrule
\end{tabular}
\end{table}
\vspace{-3mm}

\subsection{Ablation Study}
We conduct ablation studies on the turbo variant on Wan2.1-T2V-14B to evaluate the contribution of interaction-aware clustering (IC),  SNR-guided sparsity (SG), and tail-merging strategy (TM). Here, ``Anchor'' denotes the starting configuration built upon SVG2 settings. 

As reported in Table~\ref{tab:ablation_wan}, both IC and SG improve quality over Anchor, and their combination brings further gains, indicating that they are complementary. In particular, SG controls how the sparsity budget is allocated across timesteps, whereas IC improves how this budget is used for sparse region selection. Adding TM yields the best overall trade-off, improving speedup from 1.56$\times$ to 1.61$\times$ while achieving the best quality metrics. Although TM slightly reduces sparsity, it improves speed by reducing padding overhead. Overall, all three components contribute positively to final performance.

\begin{table}[htbp]
\centering
\caption{Ablation study on Wan2.1-T2V-14B}
\label{tab:ablation_wan}
\small
\setlength{\tabcolsep}{4pt}
\begin{tabularx}{\textwidth}{>{\raggedright\arraybackslash}p{3.0cm}YYYYY}
\toprule
\textbf{Method} & \textbf{PSNR}$\uparrow$ & \textbf{SSIM}$\uparrow$ & \textbf{LPIPS}$\downarrow$ & \textbf{Time}$\downarrow$ & \textbf{Sparsity} \\
\midrule
Wan2.1-T2V-14B  & --     & --     & --     & 1890s                & --    \\
\midrule
Anchor                 & 25.894 & 0.868  & 0.101  & 1227s (1.54$\times$) & 68.7\% \\
w/ IC                  & 26.435 & 0.878  & 0.093  & 1224s (1.54$\times$) & 69.9\% \\
w/ SG                  & 27.309 & 0.890  & 0.085  & 1219s (1.55$\times$) & 69.5\% \\
w/ TM                  & 26.085 & 0.872  & 0.098  & 1223s (1.55$\times$) & 67.9\% \\
w/ IC \& SG            & 28.283 & 0.903  & 0.074  & 1213s (1.56$\times$) & 70.0\% \\
\rowcolor{blue!20}
w/ IC \& SG \& TM      & \textbf{28.372} & \textbf{0.905} & \textbf{0.072} & 1171s (1.61$\times$) & 69.0\% \\
\bottomrule
\end{tabularx}
\end{table}

\vspace{-3.5mm}

\section{Conclusion \& Limitation}
\label{sec:conclusion}

We propose iCATS, a training-free sparse attention framework accelerating DiT-based video generation through improved importance estimation, sparse masking, and hardware execution. It replaces independent clustering with query–key interaction clustering and reformulates the resulting optimization for low-cost computation. SNR-guided scheduling adapts sparsity across denoising timesteps, while hardware-friendly tail merging reduces execution overhead. Together, these designs achieve state-of-the-art performance across DiT models.

Currently, the clustering procedure in iCATS is not optimized for multi-GPU parallel execution and is left for future work. The token-level sparse masking kernel also remains to be further optimized.


\begin{ack}
This work was supported by the National Natural Science Foundation of China (Grant No. 62571174), Alibaba Group through the Alibaba Innovative Research Program, and the Advanced Computing Center of Hangzhou Normal University. We thank the anonymous area chair and reviewers for their feedback on earlier drafts, and the outstanding prior works on video generation and sparse attention.
\end{ack}

\bibliographystyle{nips}
\bibliography{ref}       

\appendix
\newpage

\setcounter{figure}{0}
\setcounter{table}{0}
\setcounter{page}{1}

\textbf{\Large Appendix}

\section{SNR-Guided Sparsity}
\label{app:density}

\begin{figure}[htbp]
    \centering
    \includegraphics[width=0.68\linewidth]{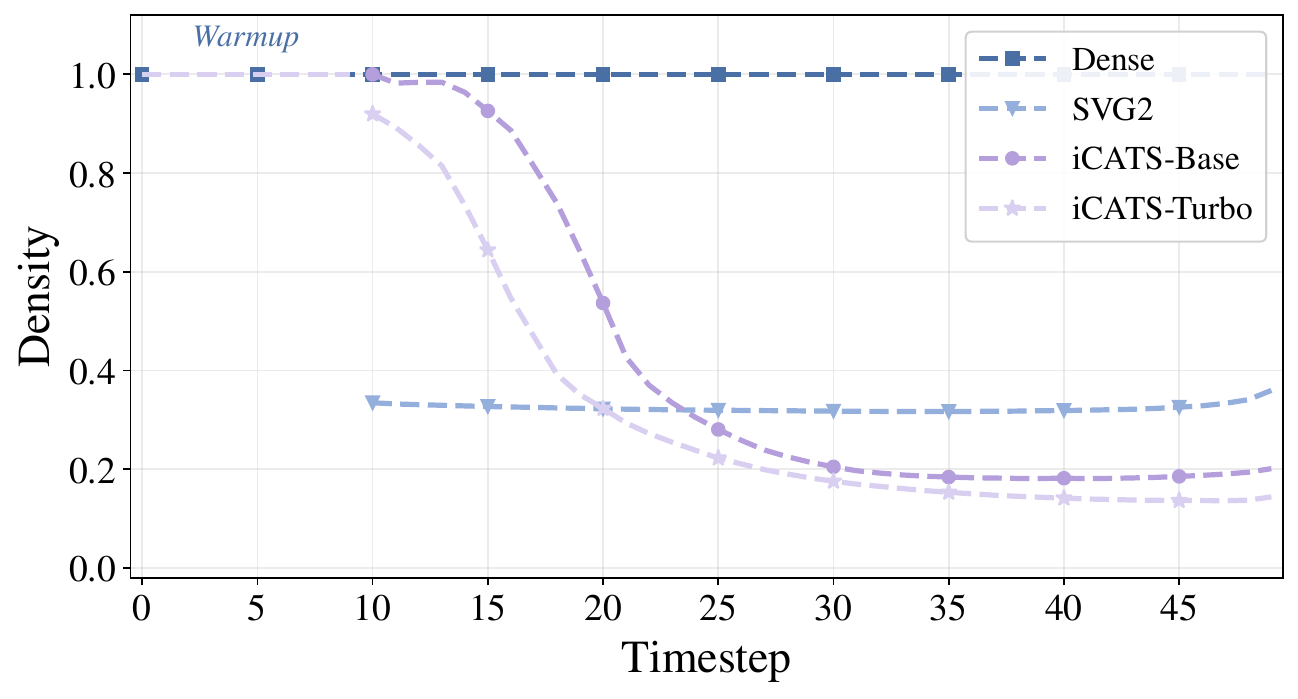}
    \caption{Density of the original full attention, SVG2, and our iCATS across timesteps on Wan2.1-T2V-14B.}
    \label{suppfig:density}
\end{figure}

Figure~\ref{suppfig:density} illustrates the efficacy of our proposed SNR-guided sparsity in iCATS. Specifically, we present the density changes across timesteps on Wan2.1-T2V-14B. 
As shown, iCATS allocates computation dynamically along the denoising trajectory: it keeps full computation during warmup, gradually reduces density in the middle stage, and stabilizes in the late stage. Compared with the fixed-density baseline SVG2 (the light blue dotted line in Figure~\ref{suppfig:density}), iCATS preserves more computation early on while using much lower density later, with iCATS-Turbo being more aggressive.

\section{Detailed Experimental Settings}
\label{app:settings}

This appendix provides the detailed settings for all compared methods.

\textbf{Compared methods}. 
We compare iCATS with two state-of-the-art sparse attention baselines, SVG and SVG2. On HunyuanVideo-T2V-13B, Wan2.1-T2V-14B and Wan2.1-I2V-14B, all baseline methods follow their official open-source implementations and default settings, including warmup timesteps, warmup layers, and the number of clustering iterations.

\textbf{iCATS variants}. 
We evaluate two variants of iCATS, Base and Turbo, designed to match the speed or sparsity of prior methods under different models. For fair comparison, the warmup timesteps, warmup layers, and clustering iterations of iCATS follow the baseline configuration of each model, as summarized in Table~\ref{tab:warmup_config}.
\vspace{-5pt}

\begin{table}[H]
\centering
\caption{Warmup and clustering configurations for each model.}
\label{tab:warmup_config}
\small
\renewcommand{\arraystretch}{1.18}
\setlength{\tabcolsep}{8pt}
\begin{tabular}{lccc}
\toprule
\textbf{Model} & \textbf{Warmup timesteps} & \textbf{Warmup layers} & \textbf{Cluster iters (first/later)} \\
\midrule
HunyuanVideo-T2V-13B & 5/50  & 1/60 & 50/2 \\
Wan2.1-T2V-14B       & 10/50 & 1/40 & 50/2 \\
Wan2.1-I2V-14B       & 14/40 & 1/40 & 50/2 \\
\bottomrule
\end{tabular}
\end{table}
\vspace{-5pt}

For \textbf{HunyuanVideo-T2V-13B}, both iCATS variants use 500 query clusters and 1000 key clusters. The Base variant uses \(p_{\min}=0.7\) and \(\tau=0.98\), while the Turbo variant uses \(p_{\min}=0.65\) and \(\tau=0.97\).

For \textbf{Wan2.1-T2V-14B}, both iCATS variants use 400 query clusters and 1000 key clusters. The Base variant uses \(p_{\min}=0.7\) and \(\tau=0.999\), while the Turbo variant uses \(p_{\min}=0.65\) and \(\tau=0.97\).

For \textbf{Wan2.1-I2V-14B}, both iCATS variants use 400 query clusters and 1000 key clusters. The Base variant uses \(p_{\min}=0.8\) and \(\tau=0.999\), while the Turbo variant uses \(p_{\min}=0.75\) and \(\tau=0.98\).

These settings are chosen to provide comparable speed or sparsity to prior sparse attention methods.

\vspace{-5pt}

\section{Additional VBench Results}
\label{app:vbench}

Table~\ref{tab:vbench_results} reports detailed VBench results for all compared methods across all models. Specifically, we report subject consistency, background consistency, motion smoothness, aesthetic quality, imaging quality, and the averaged VBench score. These fine-grained results complement the average scores reported in the main paper and provide a clearer view of quality preservation under sparse attention.

Overall, iCATS remains highly competitive with dense attention and strong sparse baselines across all models. On HunyuanVideo-T2V-13B, both iCATS-Base and iCATS-Turbo achieve an average VBench score of 0.844, exactly matching dense attention. More specifically, both variants match dense attention in subject consistency, background consistency, motion smoothness, and imaging quality. Compared with existing sparse baselines, iCATS further achieves the strongest subject consistency while remaining tied for the best background consistency and motion smoothness. These results indicate that the proposed sparse attention design preserves identity consistency, scene coherence, temporal stability, and visual fidelity without measurable degradation on this model.

On Wan2.1-T2V-14B, iCATS-Base achieves an average VBench score of 0.838, outperforming SVG (0.836) and slightly surpassing SVG2 (0.837), while iCATS-Turbo matches SVG2 at 0.837. In addition, iCATS remains competitive in subject consistency and background consistency, preserves motion smoothness close to other sparse methods, and achieves the strongest imaging quality among the sparse variants. These results suggest that the proposed sparse attention design preserves both structural coherence and visual fidelity well, while still enabling substantial efficiency gains.

On Wan2.1-I2V-14B, iCATS likewise remains highly competitive with dense attention. Both iCATS-Base and iCATS-Turbo achieve an average VBench score of 0.846, exceeding SVG (0.844) and SVG2 (0.845) while staying very close to dense attention (0.847). Looking at the detailed dimensions, iCATS-Base matches dense attention in imaging quality, iCATS-Turbo matches dense attention in background consistency, and both variants preserve the same motion smoothness as dense attention. Meanwhile, subject consistency and aesthetic quality also remain competitive with the strongest sparse baselines, showing that iCATS generalizes well beyond text-to-video generation and can maintain strong appearance consistency, scene coherence, motion quality, and perceptual fidelity in the image-to-video setting.

\begin{table}[H]
\centering
\caption{Detailed VBench comparison with existing methods on benchmark models}
\label{tab:vbench_results}
\small
\setlength{\tabcolsep}{5pt}
\renewcommand{\arraystretch}{1.18}
\resizebox{\textwidth}{!}{%
\begin{tabular}{lcccccc}
\toprule
\textbf{Method}
  & \textbf{Sub.Cons.}$\uparrow$
  & \textbf{Back.Cons.}$\uparrow$
  & \textbf{Mot.Smooth}$\uparrow$
  & \textbf{Aes.Qual.}$\uparrow$
  & \textbf{Img.Qual.}$\uparrow$
  & \textbf{VBench (Avg.)}$\uparrow$ \\
\midrule
HunyuanVideo-T2V-13B (Dense)
  & 0.940 & 0.959 & 0.994 & 0.628 & 0.698 & 0.844 \\
\midrule
SVG
  & 0.938 & \textbf{0.959} & \textbf{0.994} & \textbf{0.632} & 0.693 & 0.843 \\
SVG2
  & 0.939 & \textbf{0.959} & 0.993 & 0.627 & 0.695 & 0.842 \\
\rowcolor{blue!10}
iCATS-Base
  & \textbf{0.940} & \textbf{0.959} & \textbf{0.994} & 0.627 & \textbf{0.698} & \textbf{0.844} \\
\rowcolor{blue!20}
iCATS-Turbo
  & \textbf{0.940} & \textbf{0.959} & \textbf{0.994} & 0.627 & \textbf{0.698} & \textbf{0.844} \\
\midrule
Wan2.1-T2V-14B (Dense)
  & 0.934 & 0.955 & 0.987 & 0.630 & 0.689 & 0.839 \\
\midrule
SVG
  & 0.929 & 0.953 & \textbf{0.987} & 0.629 & 0.686 & 0.836 \\
SVG2
  & 0.931 & \textbf{0.954} & 0.986 & \textbf{0.630} & 0.684 & 0.837 \\
\rowcolor{blue!10}
iCATS-Base
  & \textbf{0.932} & \textbf{0.954} & 0.986 & 0.629 & \textbf{0.687} & \textbf{0.838} \\
\rowcolor{blue!20}
iCATS-Turbo
  & 0.931 & 0.953 & 0.986 & 0.629 & \textbf{0.687} & 0.837 \\
\midrule
Wan2.1-I2V-14B (Dense)
  & 0.958 & 0.958 & 0.986 & 0.618 & 0.716 & 0.847 \\
\midrule
SVG
  & 0.953 & 0.954 & \textbf{0.986} & 0.612 & 0.715 & 0.844 \\
SVG2
  & \textbf{0.956} & 0.955 & \textbf{0.986} & 0.614 & 0.715 & 0.845 \\
\rowcolor{blue!10}
iCATS-Base
  & \textbf{0.956} & 0.957 & \textbf{0.986} & 0.615 & \textbf{0.716} & \textbf{0.846} \\
\rowcolor{blue!20}
iCATS-Turbo
  & 0.955 & \textbf{0.958} & \textbf{0.986} & \textbf{0.616} & 0.715 & \textbf{0.846} \\
\bottomrule
\end{tabular}%
}
\end{table}

\section{Attention Fidelity Breakdown}
\label{app:fidelity_breakdown}

Table~\ref{tab:cluster_fidelity_breakdown} breaks down the attention fidelity diagnostics in Table~\ref{tab:cluster_fidelity} of the main paper by timestep, layer, and head. Interaction-aware clustering consistently outperforms independent Euclidean clustering on both metrics across all sampled points.

\begin{table}[H]
\centering
\caption{Breakdown of attention fidelity on Wan2.1-T2V-14B by timestep, layer, and head. Each entry is reported as ``interaction-aware (ours) / Euclidean''; MSE values are in units of $10^{-4}$.}
\label{tab:cluster_fidelity_breakdown}
\small
\renewcommand{\arraystretch}{1.1}
\setlength{\tabcolsep}{5pt}
\begin{tabular}{lccccc}
\toprule
\textbf{Timestep} $t$ & 952 & 882 & 768 & 555 & 92 \\
\midrule
Top-$p$ overlap$\uparrow$ & 0.902 / 0.891 & 0.914 / 0.896 & 0.915 / 0.896 & 0.916 / 0.897 & 0.917 / 0.901 \\
Output MSE$\downarrow$    & 8.91 / 11.75  & 6.33 / 9.52   & 5.41 / 8.27   & 4.71 / 7.48   & 3.40 / 7.91 \\
\midrule
\textbf{Layer} & 1 & 10 & 20 & 30 & 39 \\
\midrule
Top-$p$ overlap$\uparrow$ & 0.947 / 0.941 & 0.879 / 0.863 & 0.881 / 0.863 & 0.904 / 0.880 & 0.952 / 0.935 \\
Output MSE$\downarrow$    & 1.36 / 4.19   & 18.42 / 20.22 & 4.76 / 5.59   & 2.68 / 10.74  & 1.55 / 4.19 \\
\midrule
\textbf{Head} & 0 & 10 & 20 & 30 & -- \\
\midrule
Top-$p$ overlap$\uparrow$ & 0.933 / 0.919 & 0.920 / 0.910 & 0.903 / 0.892 & 0.894 / 0.866 & -- \\
Output MSE$\downarrow$    & 5.25 / 8.54   & 4.81 / 5.04   & 7.04 / 7.28   & 5.91 / 15.08  & -- \\
\bottomrule
\end{tabular}
\end{table}




\section{Tail-Merging Strategy}
\label{app:merge}

\subsection{Algorithm Overview}
\label{app:algo_over}
\vspace{-5pt}

\begin{algorithm}[H]
\small
\caption{Proposed Tail-merging Strategy}
\label{alg:query_cluster_merge}
\begin{algorithmic}[1]

\Require
  \Statex $\mathbf{L}\in\mathbb{Z}^{BH\times N}$: query-cluster labels;\;
          $\mathbf{s}\in\mathbb{Z}^{BH\times C}$: cluster sizes;\;
          $C$: number of query-clusters
  \Statex $\mathbf{P}\in\{0,1\}^{BH\times C\times W}$: key-cluster mask per cluster;\;
          $W$: number of key-clusters;\;
          $b$: block stride
\Ensure
  \Statex $\mathbf{L}'\in\mathbb{Z}^{BH\times N}$: updated labels;\;
          $\mathbf{s}'\in\mathbb{Z}^{BH\times C'}$: merged sizes ($C'\leq C$);\;
          $\mathbf{M}'\in\{0,1\}^{BH\times C'\times W}$: merged masks

\State $\mathbf{P}_{\mathrm{pack}} \leftarrow \mathrm{PackToInt64}(\mathbf{P})$;\;
       $(\mathbf{P}^*, \pi) \leftarrow \mathrm{Unique}([\mathrm{head\_id},\,\mathbf{P}_{\mathrm{pack}}])$
\Comment{merge clusters with identical key-cluster masks within each head}

\If{duplicate masks exist}
    \State $\mathbf{s}^* \leftarrow \mathrm{IndexAdd}(\pi,\,\mathbf{s})$;\;
           $\mathbf{L}^* \leftarrow \mathrm{Remap}(\mathbf{L},\,\pi)$
    \Comment{accumulate sizes of merged clusters; rewrite token labels via $\pi$}
\Else
    \State $\mathbf{L}^*,\,\mathbf{s}^*,\,\mathbf{P}^* \leftarrow \mathbf{L},\,\mathbf{s},\,\mathbf{P}_{\mathrm{pack}}$
\EndIf

\State $\boldsymbol{\tau} \leftarrow \mathbf{s}^* \bmod b$;\;
       $\mathbf{T} \leftarrow \mathrm{ComputeIsTail}(\mathbf{L}^*, \mathbf{s}^*, b)$
\Comment{$\boldsymbol{\tau}_i$: tail length of cluster $i$; $\mathbf{T}\in\{0,1\}^{BH\times N}$ flags tail tokens}

\If{$\exists\, \boldsymbol{\tau}_i > 0$}
    \State $\mathbf{D} \leftarrow \mathrm{HammingDist}(\mathbf{P}^*, \mathbf{P}^*)$;\; $\mathbf{D}_{ii} \leftarrow \infty$
    \State $\mathbf{D}_{ij} \leftarrow \infty\ $ if $\ \boldsymbol{\tau}_i = 0 \vee \boldsymbol{\tau}_j = 0 \vee \boldsymbol{\tau}_i + \boldsymbol{\tau}_j > b$
    \Comment{forbid self-merge, clusters without tails, and tail overflow}

    \For{$r = 1, 2$}
        \State $j^*(i) \leftarrow \arg\min_j\,\mathbf{D}_{ij}$;\;
               $\mathcal{E} \leftarrow \{(i,j) : j^*(i)=j,\, j^*(j)=i,\, i < j\}$
        \Comment{nearest neighbor by mask similarity; mutually selected pairs only}
        \If{$\mathcal{E} = \emptyset$} \textbf{break} \EndIf

        \For{$(i,j) \in \mathcal{E}$}
            \State $\mathbf{P}^*_i \leftarrow \mathbf{P}^*_i \cup \mathbf{P}^*_j$;\;
                   $\mathbf{s}^*_i \mathrel{+}= \boldsymbol{\tau}_j$;\;
                   $\mathbf{s}^*_j \mathrel{-}= \boldsymbol{\tau}_j$
            \Comment{merge masks via bitwise OR; transfer $\boldsymbol{\tau}_j$ tail tokens from $j$ to $i$}
            \State $\mathbf{L}^*_n \leftarrow i\ \ \forall n:\ \mathbf{L}^*_n{=}j \wedge \mathbf{T}_n{=}1$;\;
                   $\mathbf{D}_{j,\cdot} = \mathbf{D}_{\cdot,j} \leftarrow \infty$
            \Comment{reassign tail tokens of $j$ to $i$; invalidate $j$ in this round}
        \EndFor

        \If{$r < 2$}
            \State $\boldsymbol{\tau} \leftarrow \mathbf{s}^*\bmod b$;\;
           $\mathbf{T} \leftarrow \mathrm{ComputeIsTail}(\mathbf{L}^*,\mathbf{s}^*,b)$;\;
           recompute $\mathbf{D}$ and reapply constraints to $\mathbf{D}$
        \EndIf
    \EndFor
\EndIf

\State $\mathbf{L}' \leftarrow \mathrm{Remap}(\mathbf{L}^*,\,\mathbf{s}^* > 0)$;\;
       $\mathbf{s}' \leftarrow \mathbf{s}^*[\mathbf{s}^* > 0]$;\;
       $\mathbf{M}' \leftarrow \mathrm{Unpack}(\mathbf{P}^*[\mathbf{s}^* > 0])$
\Comment{remove empty clusters and compact indices}
\State \Return $\mathbf{L}',\; \mathbf{s}',\; \mathbf{M}'$

\end{algorithmic}
\end{algorithm}

Algorithm~\ref{alg:query_cluster_merge} summarizes the complete tail merging procedure in iCATS. Given per-token cluster assignments, cluster sizes, and binary key-cluster masks, we first pack each mask into a compact integer and, within each head, merge clusters with identical masks exactly by summing their sizes and remapping token labels. We then selectively reassign tail tokens from non-block-aligned clusters: a pair $(i, j)$ with $i<j$ is merged only if the two clusters are mutually closest in Hamming distance and their tails fit into one block, moving the tail tokens of $j$ into $i$ and merging their masks by union, which improves block utilization while largely preserving the original sparsity structure. After two rounds, empty clusters are removed and the rest are compactly reindexed within each head for subsequent sparse attention computation.

\vspace{-5pt}

\subsection{Implementation Details}
\label{app:merge_impl}

\begin{figure}[h]
  \centering
  \includegraphics[width=0.42\linewidth]{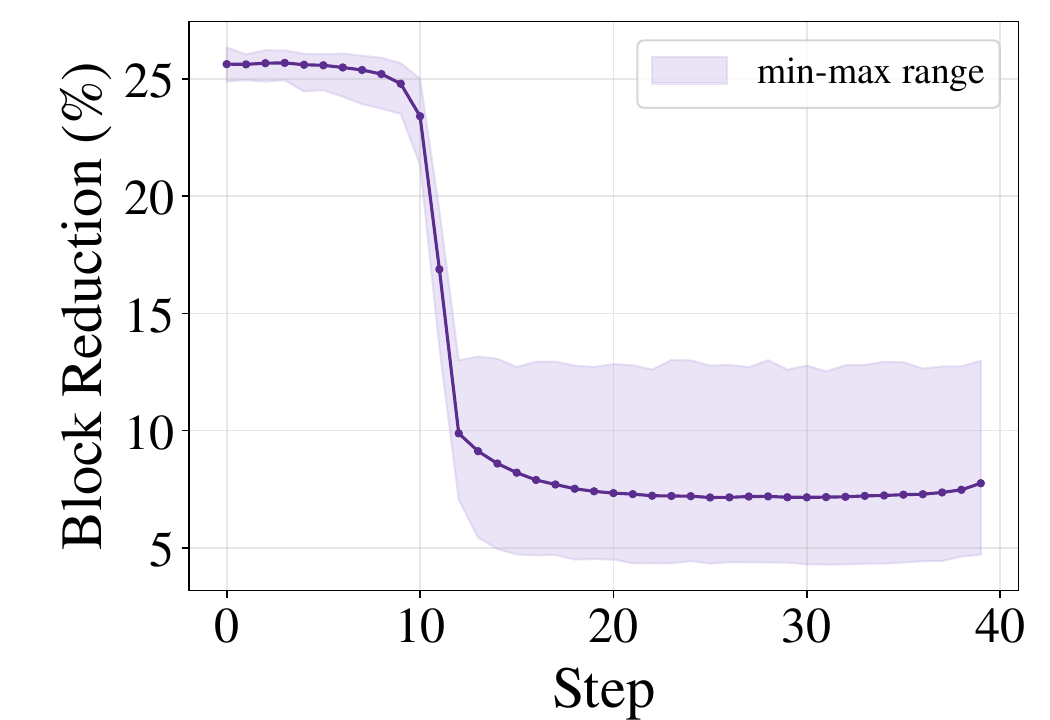}
  \caption{Relative reduction in the number of executed FlashInfer blocks on Wan2.1-T2V-14B under the iCATS-Base setting.}
  \label{suppfig:ab_block_reduction}
\end{figure}

In line with the previous cluster-based method~\cite{yang2025sparse}, in our implementation, we employ FlashInfer~\cite{ye2025flashinfer} to execute sparse attention. To evaluate the effectiveness of the tail-merging strategy, we measure the relative reduction in the number of executed blocks under the iCATS-Base setting, with SNR-guided sparsity configured by $\tau=0.999$ and $p_{\min}=0.7$. FlashInfer organizes sparse attention computation into execution blocks, e.g., with a query-side block size of 128 and a key/value-side block size of 96. As shown in Figure~\ref{suppfig:ab_block_reduction}, the average reduction across layers is about 25\% at early denoising timesteps where the sparsity level remains low. As denoising progresses and the sparsity increases according to the SNR-guided sparsity schedule (see Figure~\ref{suppfig:density}), the key/value sequence length also decreases, and thus this average reduction gradually decreases to around 7\%.
The min-max range in Figure~\ref{suppfig:ab_block_reduction} further shows the variation in block reduction ratio across sparse attention layers at each denoising timestep. 

\section{Hyperparameter Sensitivity and Transferability}
\label{app:hyperparameter_transfer}

In the main experiments, model-specific hyperparameters are chosen to match the runtime or sparsity of the corresponding baselines. To examine whether these settings can be reused across models and tasks, we directly transfer the query/key cluster counts and SNR-guided sparsity hyperparameters of HunyuanVideo-Base to Wan2.1-T2V-14B and Wan2.1-I2V-14B without retuning. Warmup timesteps, warmup layers, and clustering iterations follow each target model's baseline settings, while tail merging uses two rounds throughout. We evaluate the transferred configuration using 100 randomly sampled prompts.

Table~\ref{tab:hyperparameter_transfer} shows that the transferred configuration achieves better PSNR, SSIM, and LPIPS than SVG2 on both tasks at comparable speedup. On Wan2.1-T2V-14B, it achieves 28.656 dB PSNR, compared with 25.411 dB for SVG2, while differing from the Wan-tuned iCATS-Base configuration by only 0.404 dB. On Wan2.1-I2V-14B, it improves PSNR from 29.068 to 29.810 dB with slightly higher speedup. These results support the reuse of a shared clustering and sparsity-schedule configuration across the evaluated models and tasks.

\begin{table}[htbp]
\centering
\caption{Hyperparameter transfer from HunyuanVideo-Base to Wan2.1 without retuning.}
\label{tab:hyperparameter_transfer}
\small
\renewcommand{\arraystretch}{1.15}
\setlength{\tabcolsep}{4pt}
\begin{tabularx}{\textwidth}{>{\raggedright\arraybackslash}Xccccc}
\toprule
\textbf{Method}
& \textbf{PSNR}$\uparrow$
& \textbf{SSIM}$\uparrow$
& \textbf{LPIPS}$\downarrow$
& \textbf{Speedup}$\uparrow$
& \textbf{Sparsity} \\
\midrule
\multicolumn{6}{l}{\textbf{Wan2.1-T2V-14B}} \\
\midrule
SVG2
& 25.411 & 0.845 & 0.101 & 1.54$\times$ & 68.7\% \\
iCATS-Base (Wan-tuned)
& 29.060 & 0.904 & 0.060 & 1.54$\times$ & 60.5\% \\
iCATS (transferred)
& 28.656 & 0.899 & 0.065 & 1.55$\times$ & 66.2\% \\
\midrule
\multicolumn{6}{l}{\textbf{Wan2.1-I2V-14B}} \\
\midrule
SVG2
& 29.068 & 0.890 & 0.067 & 1.42$\times$ & 70.1\% \\
iCATS (transferred)
& 29.810 & 0.901 & 0.063 & 1.44$\times$ & 71.5\% \\
\bottomrule
\end{tabularx}
\end{table}

We further examine hyperparameter sensitivity on Wan2.1-T2V-14B using the same prompt set. Starting from the Base setting ($\tau=0.999$, $p_{\min}=0.7$), we independently decrease $\tau$ to 0.99, 0.98, and 0.97, or decrease $p_{\min}$ to 0.6, while keeping all other settings fixed. Unlike the configuration-transfer experiment, these runs retain the Wan-specific query/key cluster counts to isolate the effects of the two sparsity-schedule hyperparameters.

As shown in Table~\ref{tab:hyperparameter_sensitivity}, decreasing either parameter increases sparsity and speedup, with a gradual reduction in reconstruction quality. Across the tested $\tau$ values, sparsity increases from 60.5\% to 66.4\%, while speedup increases from 1.54$\times$ to 1.58$\times$. All tested settings retain better PSNR, SSIM, and LPIPS than SVG2, indicating that the quality advantage is maintained across the evaluated parameter range rather than depending on a single setting.

\begin{table}[htbp]
\centering
\caption{Hyperparameter sensitivity on Wan2.1-T2V-14B. Each variant changes only the indicated parameter from iCATS-Base.}
\label{tab:hyperparameter_sensitivity}
\small
\renewcommand{\arraystretch}{1.15}
\setlength{\tabcolsep}{4pt}
\begin{tabularx}{\textwidth}{>{\raggedright\arraybackslash}Xccccc}
\toprule
\textbf{Setting}
& \textbf{PSNR}$\uparrow$
& \textbf{SSIM}$\uparrow$
& \textbf{LPIPS}$\downarrow$
& \textbf{Speedup}$\uparrow$
& \textbf{Sparsity} \\
\midrule
SVG2
& 25.411 & 0.845 & 0.101 & 1.54$\times$ & 68.7\% \\
iCATS-Base
& 29.060 & 0.904 & 0.060 & 1.54$\times$ & 60.5\% \\
\midrule
$\tau=0.99$
& 28.716 & 0.900 & 0.063 & 1.55$\times$ & 63.4\% \\
$\tau=0.98$
& 28.537 & 0.897 & 0.065 & 1.56$\times$ & 65.2\% \\
$\tau=0.97$
& 28.392 & 0.895 & 0.067 & 1.58$\times$ & 66.4\% \\
$p_{\min}=0.6$
& 28.576 & 0.896 & 0.066 & 1.57$\times$ & 63.8\% \\
\bottomrule
\end{tabularx}
\end{table}

\section{Runtime Breakdown}
\label{app:runtime}

We profile iCATS-Turbo on Wan2.1-T2V-14B over 10 randomly sampled prompts using NVIDIA H100 GPUs and CUDA events. We use 50 denoising steps, 40 layers, 10 dense warmup steps, one dense warmup layer, and classifier-free guidance, with approximately 70\% average sparsity over sparse timesteps. Timings cover denoising only, excluding model loading, text encoding, and VAE decoding.

\begin{table}[htbp]
    \centering
    \caption{Runtime breakdown of iCATS-Turbo.}
    \label{tab:runtime_breakdown}
    \small
    \renewcommand{\arraystretch}{1.15}
    \setlength{\tabcolsep}{4pt}
    \begin{tabular}{lr}
        \toprule
        \textbf{Component} & \textbf{Time (s)} \\
        \midrule
        Dense attention          & 286.5 \\
        Sparse attention kernel  & 263.3 \\
        Clustering               & 78.5  \\
        Token permutation        & 11.2  \\
        Tail merging             & 10.6  \\
        Sparse mask construction & 3.8   \\
        Others (proj., QK-norm, RoPE, FFN, cross-attn., norm, etc.) & 507.1 \\
        \midrule
        Total                    & 1161.0 \\
        \bottomrule
    \end{tabular}
\end{table}
Clustering accounts for 6.8\% of the total denoising time. Following prior methods, we use 50 initialization iterations and two refinement iterations at later timesteps, warm-started from previous centroids. Other DiT operations include projections, QK normalization, RoPE, FFNs, cross-attention, and normalization.

\section{Additional Visualizations}
\label{app:visual}

We provide additional visualization comparisons between iCATS and dense attention on Wan2.1-T2V-14B. These examples demonstrate that iCATS preserves high visual fidelity and maintains generation quality that is visually close to the original dense attention across diverse scenes and motion patterns.

\begin{figure}[htbp]
    \centering
    \includegraphics[width=0.98\linewidth]{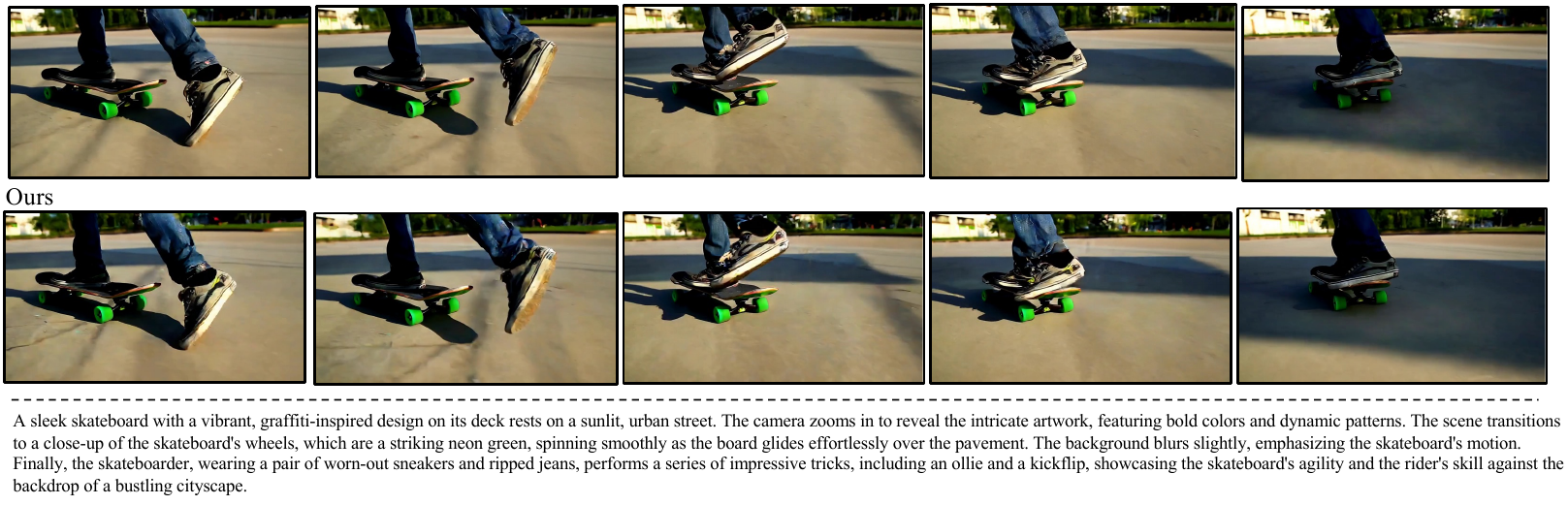}\par\vspace{12pt}
    \includegraphics[width=0.98\linewidth]{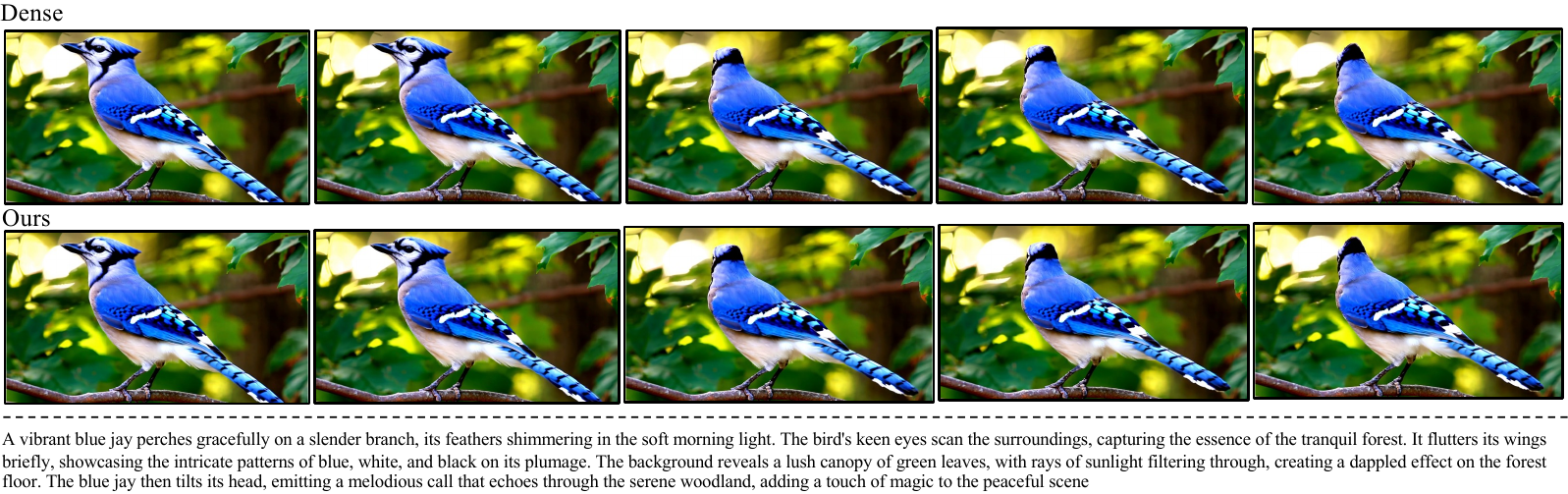}\par\vspace{12pt}
    \includegraphics[width=0.98\linewidth]{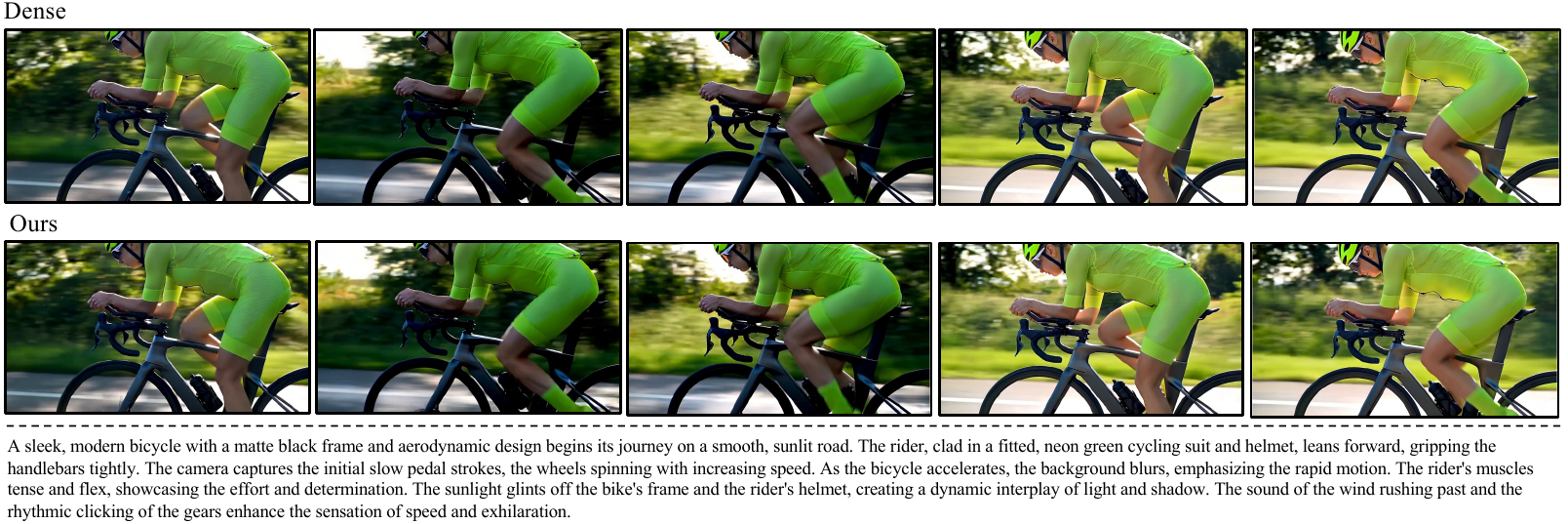}\par\vspace{12pt}
    \includegraphics[width=0.98\linewidth]{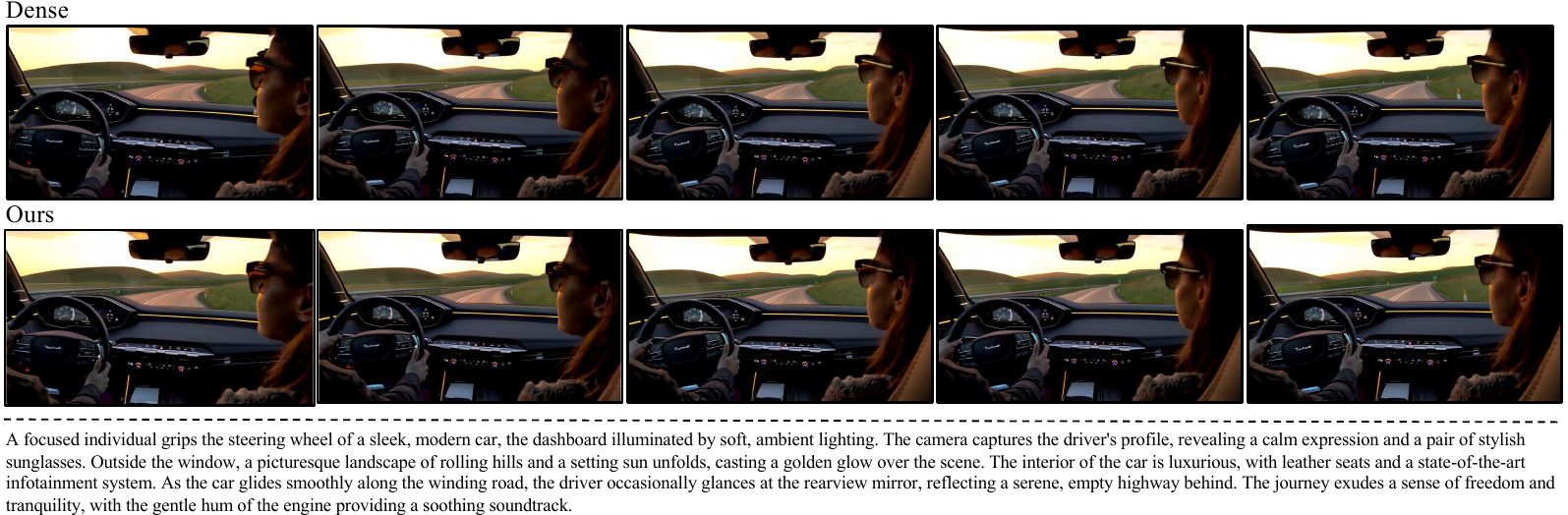}
    \caption{Visualization comparison between iCATS and dense attention on Wan2.1-T2V-14B.}
    \label{fig:icats_vs_dense_wan_t2v}
\end{figure}

\clearpage
\begingroup
\makeatletter
\setlength{\@fptop}{0pt}
\setlength{\@fpbot}{0pt plus 1fil}
\makeatother

\begin{figure}[p]
    \ContinuedFloat
    \centering
    \includegraphics[width=0.98\linewidth]{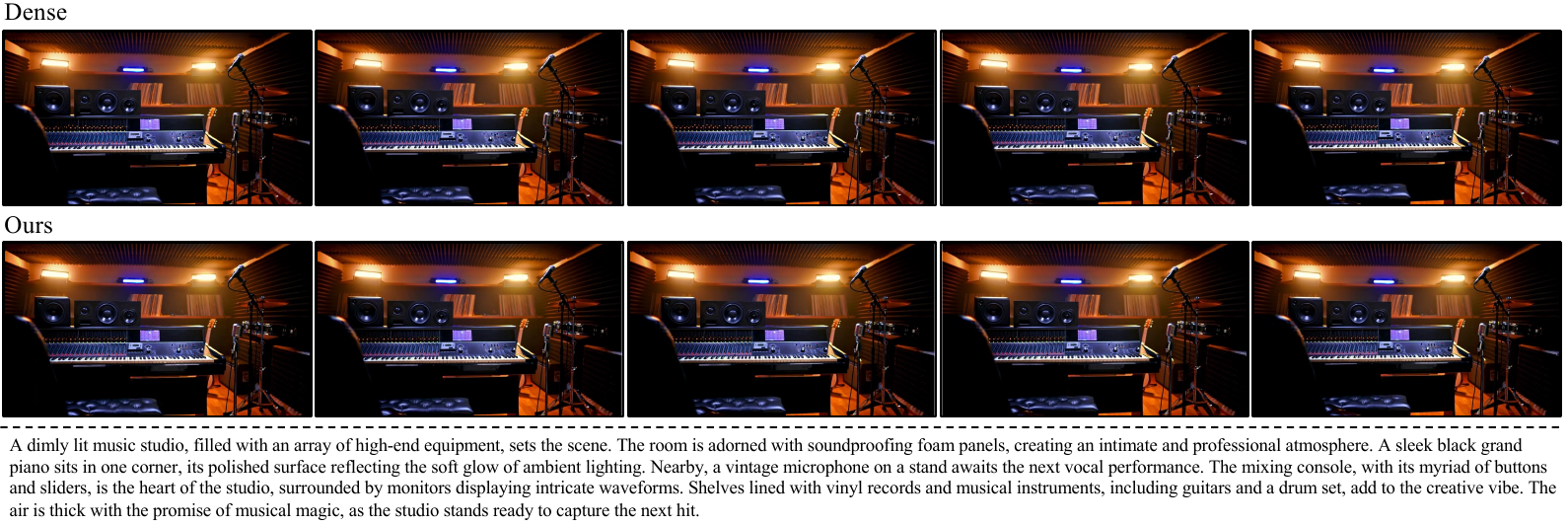}\par\vspace{12pt}
    \includegraphics[width=0.98\linewidth]{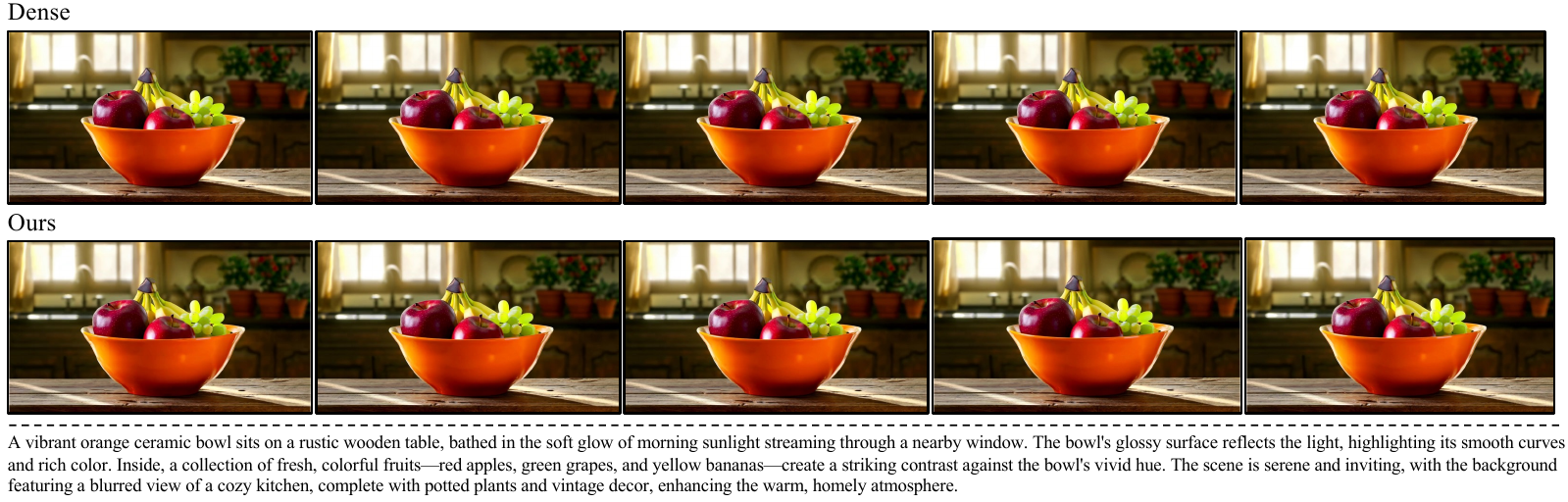}
    \caption[]{\textit{Figure~\ref{fig:icats_vs_dense_wan_t2v} continued.}}
\end{figure}

\clearpage
\endgroup
\clearpage

\end{document}